\documentclass{article}

\usepackage{microtype}
\usepackage{graphicx}
\usepackage{booktabs} 

\usepackage{hyperref}

\usepackage[accepted]{icml2025}

\usepackage{amsmath}
\usepackage{amssymb}
\usepackage{mathtools}
\usepackage{amsthm}

\usepackage[capitalize,noabbrev]{cleveref}

\usepackage{hyperref}
\usepackage{url}
\usepackage{booktabs}       
\usepackage{amsfonts}       
\usepackage{nicefrac}       
\usepackage{microtype}      

\usepackage{amsmath}
\usepackage{amssymb}
\usepackage{mathtools}
\usepackage{amsthm}

\usepackage{times}
\usepackage{soul}
\usepackage{url}
\usepackage[utf8]{inputenc}
\usepackage[small]{caption}
\usepackage{graphicx}
\usepackage{amsmath}
\usepackage{amsthm}
\usepackage{algorithm}
\usepackage{algorithmic}
\usepackage{url}
\usepackage{booktabs}       
\usepackage{amsfonts}       
\usepackage{nicefrac}       
\usepackage{microtype}      
\usepackage{xcolor}         
\usepackage{booktabs} 
\usepackage{times}
\usepackage{epsfig}
\usepackage{amssymb}
\usepackage{pifont}
\usepackage{kotex}
\usepackage{enumitem}
\usepackage{verbatim}
\usepackage{bm}
\usepackage{makecell,multirow, tabularx}
\usepackage{balance}
\usepackage{upquote}
\usepackage{appendix}
\usepackage{nicefrac}
\usepackage{physics}
\usepackage{stmaryrd}
\usepackage{pifont}
\usepackage{wrapfig} 
\setlist{nosep}
\usepackage{graphicx}
\usepackage{subcaption}
\usepackage{xspace}
\newcommand\eg{\textit{e.g.}\xspace}

\usepackage{color, colortbl}

\theoremstyle{plain}

\theoremstyle{definition}

\theoremstyle{remark}

\usepackage[textsize=tiny]{todonotes}

\icmltitlerunning{A Multiagent Approach and Dual Evaluation Metrics for Factuality and Coverage}

\begin{document}

\twocolumn[
\icmltitle{Toward Robust Hyper-Detailed Image Captioning: A Multiagent Approach and Dual Evaluation Metrics for Factuality and Coverage}



\icmlsetsymbol{intern}{*}

\begin{icmlauthorlist}
\icmlauthor{Saehyung Lee}{snu,intern}
\icmlauthor{Seunghyun Yoon}{adobe}
\icmlauthor{Trung Bui}{adobe}
\icmlauthor{Jing Shi}{adobe}
\icmlauthor{Sungroh Yoon}{snu,snuu}
\end{icmlauthorlist}

\icmlaffiliation{snu}{Department of Electrical and Computer Engineering, Seoul National University}
\icmlaffiliation{snuu}{Interdisciplinary Program in Artificial Intelligence, Seoul National University}
\icmlaffiliation{adobe}{Adobe Research}

\icmlcorrespondingauthor{Sungroh Yoon}{sryoon@snu.ac.kr}

\icmlkeywords{Machine Learning, ICML}

\vskip 0.3in
]



\printAffiliationsAndNotice{\intern}  

\begin{abstract}
    Multimodal large language models (MLLMs) excel at generating highly detailed captions but often produce hallucinations. Our analysis reveals that existing hallucination detection methods struggle with detailed captions. We attribute this to the increasing reliance of MLLMs on their generated text, rather than the input image, as the sequence length grows. To address this issue, we propose
    a multiagent approach that leverages LLM-MLLM collaboration to correct given captions.
    Additionally, we introduce an evaluation framework and a benchmark dataset to facilitate the systematic analysis of detailed captions. Our experiments demonstrate that
    our proposed evaluation method better aligns with human judgments of factuality than existing metrics and that
    existing approaches to improve the MLLM factuality may fall short in hyper-detailed image captioning tasks. In contrast, our proposed method significantly enhances the factual accuracy of captions, even improving those generated by GPT-4V. Finally, we highlight a limitation of VQA-centric benchmarking by demonstrating that an MLLM's performance on VQA benchmarks may not correlate with its ability to generate detailed image captions. Our code and data are available at \url{https://github.com/adobe-research/CapMAS}.
\end{abstract}
\section{Introduction}
Numerous image captioning methods utilizing deep neural networks (DNNs) have been proposed \citep{show_and_tell,show_attend_and_tell}. However, they are generally limited to generating short captions, which constrains their broader application in real-world scenarios. For instance, in cases such as assistance for visually impaired individuals, where it is necessary to provide highly detailed descriptions of the scene in front of the user, these methods may not be suitable.

Following the recent success of large language models (LLMs) \citep{gpt3}, there have been attempts to use text and information from other modalities as input to LLMs. Notably, many studies have explored multimodal large language models (MLLMs) that incorporate visual information \citep{blip2,instructblip,llava}. These models have demonstrated significantly superior performance compared to traditional models in tasks such as visual question answering (VQA) and captioning \citep{llavanext}. In particular, MLLMs, leveraging the advanced language capabilities of LLMs, are able to generate much longer and more detailed captions than conventional captioning models. However, these generated captions frequently contain inaccurate information, including descriptions of objects that are not present in the input image \citep{vcd}. Such hallucination problems hinder the practical application of MLLMs in real-world settings.

Three major approaches have been recently proposed to improve the factuality of MLLMs: (\romannumeral 1) Decoding-based methods \citep{vcd} reduce the probabilities of hallucination-related tokens during the model's decoding process without requiring additional training; (\romannumeral 2) Training-based methods \citep{robust_instruction_tuning} further train the models on curated multimodal datasets to ensure they generate only accurate responses; (\romannumeral 3) Corrector-based methods \citep{lure} employ a corrector model that detects and either removes or revises hallucinations present in the model's responses.

In this paper, \textbf{we propose Caption factuality enhancing MultiAgent System} (CapMAS), a
multiagent approach to correct hyper-detailed image captions. Unlike existing corrector-based approaches that require training a corrector \citep{volcano}, CapMAS improves the factuality of detailed image captions by leveraging the collaboration between an LLM and MLLM, without the need for additional training. Moreover, unlike methods that target specific types of hallucinations \citep{pope,lure}, our approach does not pre-define the hallucination types, allowing it to address a broader range of issues. The method proceeds as follows: (\romannumeral 1) an LLM decomposes a given detailed caption into atomic propositions; (\romannumeral 2) an MLLM verifies the truthfulness of each atomic proposition based on the image; and (\romannumeral 3) the LLM revises the caption accordingly. Our design is particularly motivated by the observation that, as the length of a model's response increases, hallucinations generated later in the sequence become more difficult for existing methods \citep{self_consistency,lure} to detect.

Evaluating the factuality of detailed captions is not straightforward. Through experiments, we demonstrate that conventional caption evaluation metrics such as BLEU \citep{bleu}, ROUGE \citep{rouge}, METEOR \citep{meteor}, and CIDEr \citep{cider}, as well as recently proposed methods \citep{clipscore,aloha}, fail to accurately assess the factuality of detailed captions. To address this issue, \textbf{we propose a GPT-based method for factuality evaluation and validate its effectiveness through experiments that include human evaluations}. Even if a caption contains factual information, however, it may still be considered inadequate if it does not sufficiently capture the visual information. To measure the coverage of captions, \textbf{we construct a detailed VQA dataset through a collaboration between humans and an AI agent} \citep{gpt4}. If a caption fully encapsulates the information of a given image, questions about the image should be answerable accurately using only the caption, without referencing the image itself.

Our experiments surprisingly reveal that methods designed to improve the factuality of MLLMs, which have proven effective in tasks like VQA
\citep{opera}, may be ineffective for hyper-detailed image captioning tasks that require longer responses. In contrast, CapMAS significantly enhances the factuality of captions and can be applied in a plug-and-play manner to any captioning model; notably, this improvement extends to captions generated by the state-of-the-art closed model, GPT-4V \citep{gpt4}. Finally, we highlight an issue with the current VQA-centric benchmarking \citep{opencompass} by showing that an MLLM's performance on VQA benchmarks may not correlate with its ability to generate detailed image captions.

\section{Related Work}
\paragraph{Multimodal large language models.}
LLMs that process inputs from multiple modalities, including text and other types of data, are referred to as multimodal LLMs \citep{mllm_survey}. Among these, \textbf{LLMs that handle visual input have been the most actively researched, and the MLLMs discussed in this paper are focused on this category}. Research on these models primarily explores methods for fusing the output of an independent vision encoder into the input of an LLM. The BLIP models \citep{blip2,instructblip} align the frozen vision encoder and LLM using a lightweight transformer \citep{transformer} called Q-Former. The trainable input tokens of the Q-Former interact with the output tokens from the vision encoder through cross-attention, transforming them into input tokens for the LLM.
The LLaVA models \citep{llava,llavanext} use a simple MLP connector to align the vision encoder with the LLM. All output tokens from the vision encoder, passed through the MLP connector, are used as input to the LLM. The vision encoder's parameters remain fixed during the training of the MLP connector and the LLM.
Unlike existing MLLMs, the InternVL models \citep{internvl,internvl_1_5} have demonstrated the effectiveness of increasing the size of both the vision encoder and the vision-language connector. They utilize a 6-billion parameter vision encoder and an 8-billion parameter vision-language connector. The connector is obtained by fine-tuning the pre-trained multilingual LLaMA \citep{qllama}. Despite the many advancements in open-source MLLMs, closed-source MLLMs such as GPT-4V or GPT-4o\footnote{\url{https://openai.com/index/hello-gpt-4o/}} still outperform them significantly. As a result, these GPT models represent the upper bound performance in benchmarks and are commonly used to evaluate MLLMs \citep{aloha}. In our work, we demonstrate that captions generated by GPT-4V can be improved using our method, and we use GPT-4o to evaluate captions.
\paragraph{MLLM hallucinations and mitigation strategies.}
MLLMs sometimes generate inaccurate responses. For example, they may incorrectly describe the characteristics of objects in an input image, misrepresent relationships between objects, or even describe objects that do not exist. To mitigate these hallucination problems, decoding-based methods apply penalties to the probabilities of tokens that are likely to be hallucinations during the decoding process. For instance, VCD \citep{vcd} induces hallucinations using corrupted images, while OPERA \citep{opera} leverages the correlation between high attention weights assigned to a few summary tokens and hallucinations.
Training-based methods focus on exploring training data that can suppress the generation of hallucinations. \citet{robust_instruction_tuning} demonstrated that hallucinations can be alleviated by incorporating negative samples—descriptions that explicitly state the absence of certain objects in a given image—into visual instruction tuning datasets.
Corrector-based methods \citep{lure,volcano} detect, remove, and revise hallucinations present in MLLM responses by using a corrector model. This model is obtained by supervised fine-tuning a pre-trained MLLM. The corrector model then revises the initial response based on the given image.
\paragraph{Caption evaluation methods.}
\begin{figure*}[t]
{
\includegraphics[width=\textwidth]{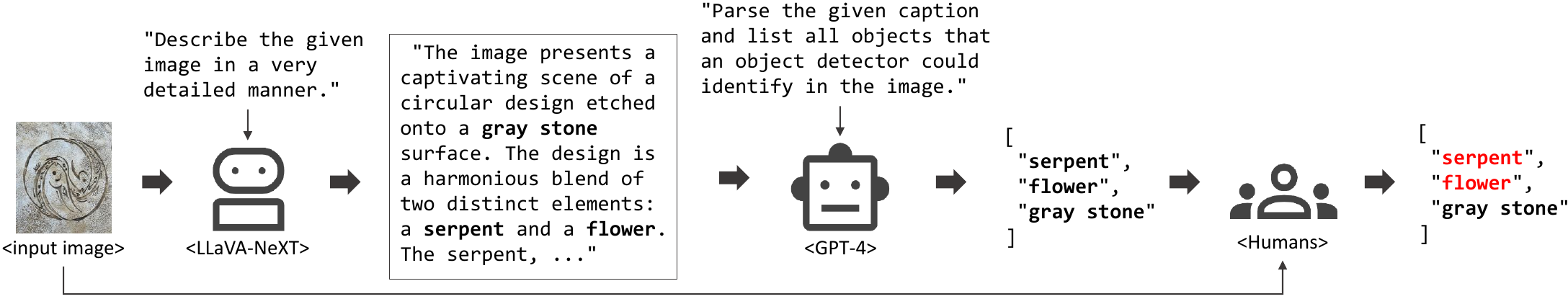}
\caption{The process of generating a data sample for evaluating hallucination detection methods in detailed image captioning tasks. Human annotators identify and label \textcolor{red}{object hallucinations} within the caption generated by LLaVA-NeXT \citep{llavanext} for an image.}
\label{fig:motivation_dataset}
}
\end{figure*}
Since short image captions are relatively easy to obtain reference captions for, we can use matching-based caption evaluation methods \citep{caption_survey} to assess them. However, for long and detailed captions generated by MLLMs, the number of reference captions required for such evaluations becomes exceedingly large. Thus, it becomes impractical to evaluate detailed captioning using traditional approaches.
\citet{clipscore} proposed CLIPScore, a reference-free evaluation method. CLIPScore measures the distance between an image and its caption within the joint representation space of CLIP \citep{clip}. Additionally, the authors introduced RefCLIPScore, which uses both the image and reference captions within that same representation space.
\citet{clair} addressed the limitations of matching-based methods by utilizing an LLM. The LLM-based metric they proposed, CLAIR, assigns scores to captions based on reference captions using an LLM.
Similarly, ALOHa \citep{aloha} detects object hallucinations by comparing generated captions to reference captions using an LLM.
\section{Method}
In this paper, we propose a multiagent-based caption correction method. Corrector-based methods typically detect and remove hallucinations within model responses. Unlike existing approaches, which require the corrector model training, our method employs collaboration between an MLLM and LLM. Moreover, in contrast to previous methods that are limited to correcting specific types of hallucinations \citep{lure}, our approach is free from such constraints.
We also propose a framework for evaluating the detailed image captioning capabilities of an MLLM. Unlike existing methods, our proposed evaluation approach allows for assessing image captioning models in terms of both factuality and coverage, evaluating each of these aspects separately.
\subsection{Motivating Observations}\label{sec:motivating_observation}
\begin{figure}
{
     \centering
     \begin{subfigure}{0.495\columnwidth}
         \includegraphics[width=\columnwidth]{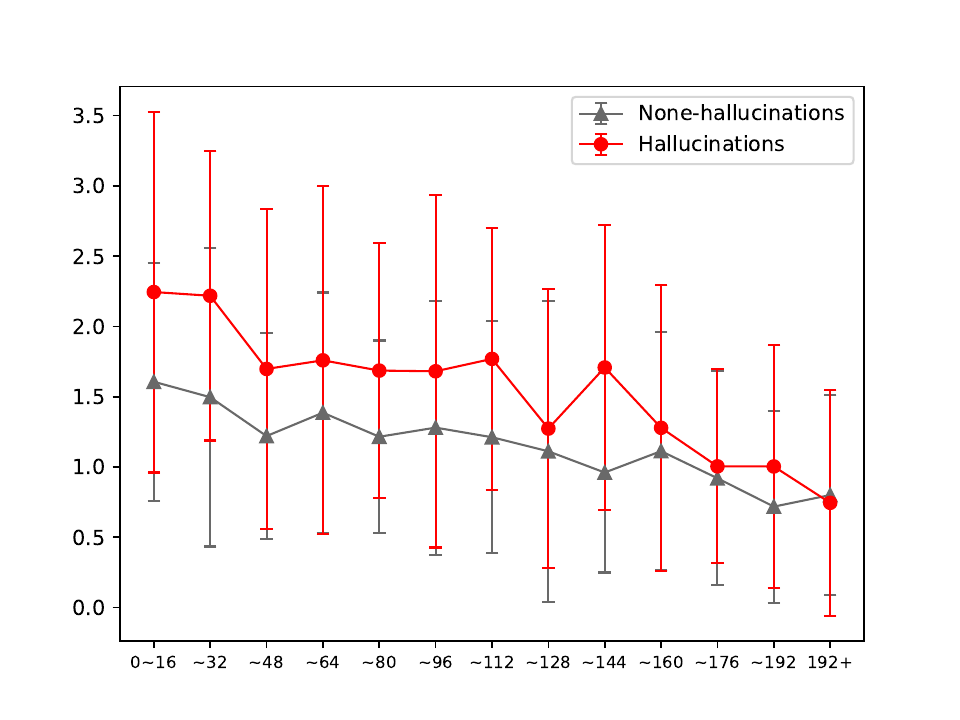}
         \caption{Confidence-Token Index}
         \label{fig:confidence}
     \end{subfigure}
     \begin{subfigure}{0.495\columnwidth}
         \centering
         \includegraphics[width=\columnwidth]{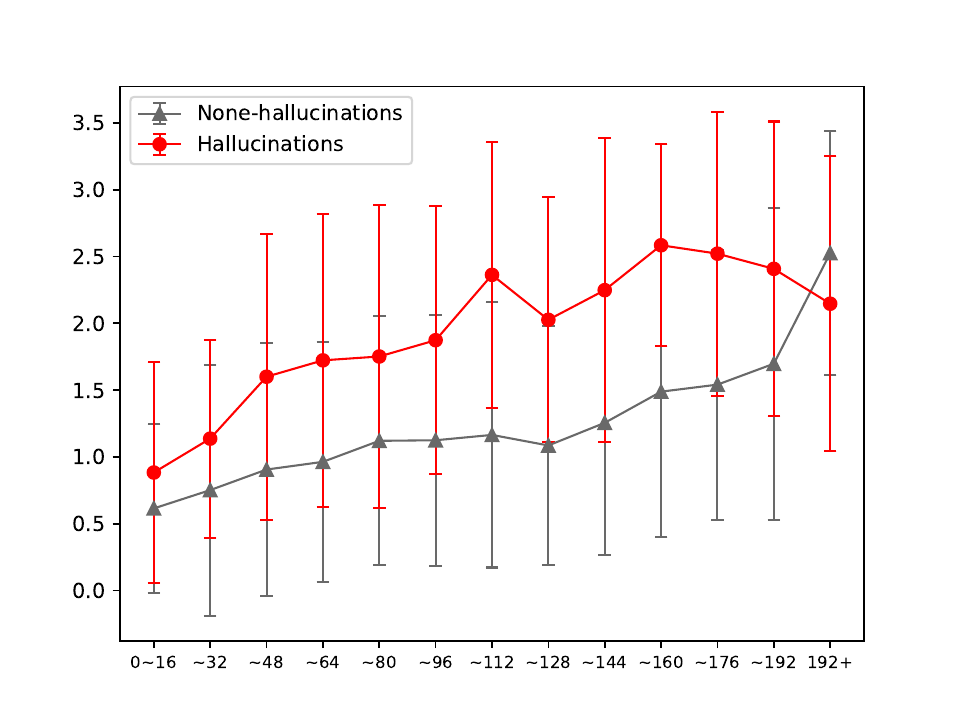}
         \caption{Consistency-Token Index}
         \label{fig:consistency}
     \end{subfigure}
        \caption{The hallucination scores of the Confidence and Consistency methods based on object positions within detailed captions. Object hallucinations near the end of the captions (192+) are undetectable by both methods.}
        \label{fig:motivation_results}
        }
\end{figure}
Here, we examine the performance of existing hallucination detection methods on tasks that require generating long responses. To facilitate these analyses, we construct a dataset as follows:
(\romannumeral 1) We prompt an MLLM with ``Describe the given image in a very detailed manner.'' and collect the model's responses for a specified image set;
(\romannumeral 2) For the convenience of our analysis, we use an LLM to identify objects that may be hallucinations;
(\romannumeral 3) Human annotators then label each parsed object as either a hallucination or not, based on the corresponding image.
We use LLaVA-NeXT \citep{llavanext} and GPT-4 as the MLLM and LLM, respectively. Figure \ref{fig:motivation_dataset} illustrates the process of constructing the dataset. To build the dataset, we use a subset of IIW-400 \citep{iiw}. We detect object hallucinations using two of the most widely adopted hallucination detection methods:
\begin{enumerate}[leftmargin=*, nolistsep]
    \item \textbf{Confidence} \citep{confidence1, lure}: This method detects hallucinations using the predicted probability $p_{\textup{obj}}$ for the object token during LLaVA-NeXT generation. For multi-token objects, the token probabilities are multiplied. The hallucination score $H_{\textup{obj}}=-\log p_{\textup{obj}}$, increases with the likelihood of hallucination.
    \item \textbf{Consistency} \citep{self_consistency, zhao2024knowing}: This method assumes that hallucinations are sensitive to decoding randomness. Using stochastic decoding, we have LLaVA-NeXT generate 40 captions per image and count the occurrence $t_{\textup{obj}}$ of each object in the dataset of Figure \ref{fig:motivation_dataset}. The hallucination score is $H_{\textup{obj}}=-\log\frac{t_{\textup{obj}}}{40}$.
\end{enumerate}
Figure \ref{fig:motivation_results} presents the hallucination scores of each method by the position of objects appearing within detailed captions. The horizontal axis of the graphs represents bins of object token indices, with larger token indices indicating positions closer to the end of the caption. The vertical axis represents the mean and standard deviation of the hallucination scores within each bin. Note that Figure \ref{fig:confidence} reflects the positions and hallucination scores during greedy decoding, while Figure \ref{fig:consistency} is derived from the average positions and hallucination scores across 40 stochastic decoding iterations. Figure \ref{fig:motivation_results} demonstrates that hallucinations generated after the 192nd token are undetectable by the Confidence and Consistency methods. Based on these results, we can infer that existing hallucination detection methods may be ineffective in detecting hallucinations in long detailed captions.

\begin{table}[]
\centering
\caption{Performance comparison of hallucination detection methods for the dataset of Figure \ref{fig:motivation_dataset}.}\label{tab:motivation}
{
\begin{tabular}{c @{\extracolsep{\fill}} cc}
Method & AUROC$\uparrow$ & FPR95$\downarrow$\\\midrule[.1em]
Confidence & 57.5 & 95.1\\
Consistency & 73.5 & 75.6\\
Object Detector & 61.5 & 95.7\\
Isolation & \textbf{81.4} & \textbf{71.7}\\
\midrule[.1em]
\end{tabular}
}
\end{table}
Our hypothesis regarding these results is that \emph{as MLLM outputs become longer, they become more strongly grounded in the text they generate rather than the given image.} In fact, our hypothesis is supported by several recent studies. For example, \citet{hypothesis1} demonstrated that as MLLM responses lengthen, the attention weights assigned to image tokens decrease, and \citet{hypothesis2} showed that MLLM responses are significantly influenced by prior dialogue. Based on this hypothesis, we test a method for determining whether each object is a hallucination by disconnecting it from its context (\textbf{Isolation}). The Isolation method involves querying the LLaVA-NeXT model with parsed objects using the prompt template, ``Is there a \{\} in the photo?'' along with the image. When the probability of the ``Yes'' token for the object query is $p_{\textup{Yes}|\textup{obj}}$, the hallucination score is defined as $H_\textup{obj} = -\log p_{\textup{Yes}|\textup{obj}}$. We compare the object hallucination detection performance of the Isolation method with that of the Confidence method, the Consistency method, and a method based on an object detector (\textbf{Object Detector}) introduced in recent studies \citep{woodpecker,vfc}. We measure their detection performance on the dataset of Figure \ref{fig:motivation_dataset} using Area Under the Receiver Operating Characteristic (AUROC) and False Positive Rate at 95\% true positive rate (FPR95). Table \ref{tab:motivation} demonstrates that the Isolation method outperforms the others. This suggests that breaking a long caption into smaller units and examining each individually can help detect hallucinations in detailed captions.

\paragraph{Comparison with existing studies.}
Actually, the concept of decomposing text into smaller units and assessing the factuality of each has been introduced in prior studies \citep{factscore,faithscore}. Here, we summarize the key differences between our study and previous research:
\begin{enumerate}[leftmargin=*, nolistsep]
    \item Unlike existing studies, which do not rigorously justify the need for decomposition, we empirically demonstrate the motivation behind our approach (Section \ref{sec:motivating_observation}).
    \item While previous studies focused solely on proposing evaluation metrics, our research advances further by introducing a system that leverages the decomposition process to generate improved image captions (Section \ref{sec:facter}).
    \item Existing evaluation metrics assess factuality using unimodal data (either text or image). In contrast, our proposed evaluation metric utilizes multimodal data (both text and image) for factuality assessment (Section \ref{sec:eval}).
    \item Previous studies focus solely on measuring the factuality of generated text. In contrast, our study proposes a method that assesses both the factuality and coverage of any given image caption (Section \ref{sec:eval}).
    \item We demonstrate that our metric correlates more strongly with human evaluations than existing metrics and is robust against their critical limitations (Section \ref{sec:human_eval}).
\end{enumerate}
\subsection{Caption Factuality Enhancing MultiAgent System}\label{sec:facter}
\begin{figure*}[t]
{
\begin{center}
\centerline{\includegraphics[width=1.0\textwidth]{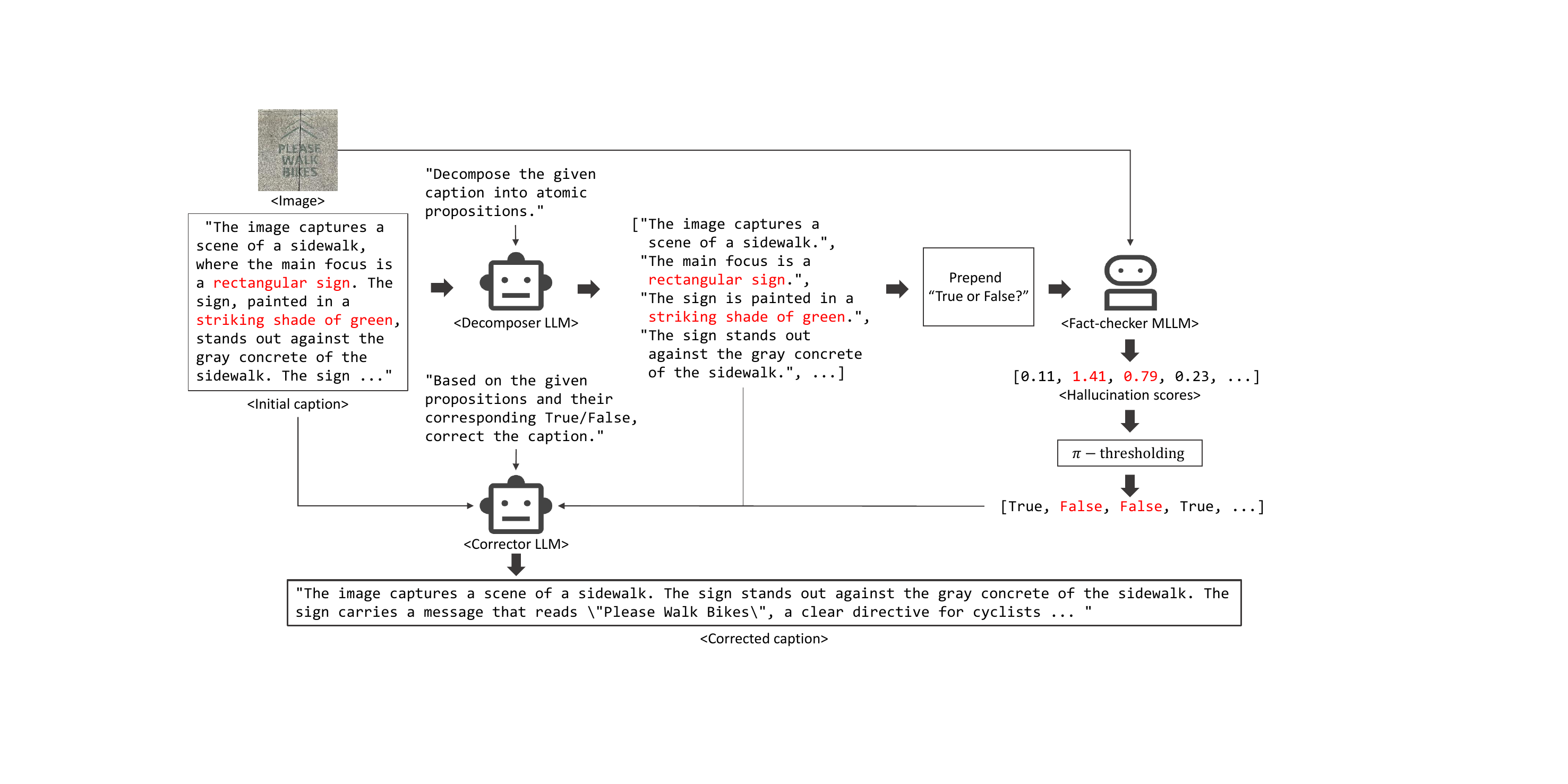}}%
\caption{
Overview of CapMAS. The decomposer LLM breaks an initial caption into atomic units. These units are converted into True/False questions and fed into the MLLM along with the image, where each unit is assigned a \textcolor{red}{hallucination} score according to Equation ($\ref{eq:facter}$). Units are classified as True or False based on the threshold $\pi$, and the corrector LLM then revises the initial caption accordingly.
}
\label{fig:method_facter}
\end{center}
}
\end{figure*}
To address various types of hallucinations comprehensively, we first decompose each detailed caption into atomic propositions using an LLM. An atomic proposition is a claim or statement that must either be true or false. For example, the caption ``A house has a red roof and a chimney'' is broken down into ``A house has a red roof'' and ``A house has a chimney.'' We use an LLM to perform this process, but we allow flexibility in cases where the results do not strictly conform to the definition of an atomic proposition. We then investigate the truth of each decomposed unit using an MLLM. Each unit is converted into a True/False question and independently fed to the MLLM. The hallucination score $H(u)$ for the unit $u$ is defined as follows:
\begin{equation}\label{eq:facter}
    -\log\left(\min\left(p\left(\textup{T}|x, Q(u)\right) - p\left(\textup{F}|x, Q(u)\right), \epsilon\right)\right)
\end{equation}
$p\left(\textup{T}\right)$ and $p\left(\textup{F}\right)$ represent the MLLM's token probabilities for the ``True'' and ``False'' tokens, respectively.
$x$ and $\epsilon$ denote the input image and a very small constant near zero. $Q(\cdot)$ is a function that converts the input text into a True/False question, which we implement by prepending ``True or False?'' to the input. Each unit is included in either the True set $\mathcal{T}$ or the False set $\mathcal{F}$, based on its hallucination score. To achieve this, we introduce a hyperparameter $\pi$, such that $\mathcal{T}=\{u|H(u)\leq\pi\}$ and $\mathcal{F}=\{u|H(u)>\pi\}$. Finally, the initial caption, along with the corresponding sets $\mathcal{T}$ and $\mathcal{F}$, is provided to an LLM, which corrects the initial caption to ensure it contains only factual information.

We name this pipeline, which improves the factuality of detailed image captions through the collaboration of a pre-trained LLM and MLLM, \textbf{Caption factuality enhancing MultiAgent System (CapMAS)}. CapMAS is training-free and can be applied in a plug-and-play manner to any captioning model. Unlike existing methods that can only address predefined types of hallucinations, CapMAS can detect and correct all hallucinations at the atomic unit level. The pipeline of CapMAS is illustrated in Figure \ref{fig:method_facter}.
\subsection{Evaluation Methods}\label{sec:eval}
Traditional caption evaluation methods rely on word matching with reference captions, suitable for short captions generated by conventional models. However, MLLMs produce longer and more detailed captions, making it impractical to obtain sufficient reference captions for accurate evaluation.
Given the enriched content of these image captions, rather than simply evaluating them as good or bad, we aim to assess them systematically by considering two key perspectives:
\begin{itemize}[leftmargin=*, nolistsep]
    \item \textbf{Factuality}: The degree to which the content of the caption is factual and free from hallucinations.
    \item \textbf{Coverage}: The extent to which the caption captures the information contained in the image.
\end{itemize}
We propose evaluation methods for detailed image captions from these two perspectives.
\paragraph{Factuality.}\begin{table}[]
\centering
\caption{Meta-evaluation results across various caption evaluation methods. DOCCI and its synthetic hallucinatory captions are used for the meta-evaluation. The highest-rated caption for each method is highlighted in \textbf{bold}. The full table is in Appendix \ref{app:metaeval_full}.}
{
\footnotesize
\addtolength{\tabcolsep}{-3.5pt}
\begin{tabular*}{1.0\columnwidth}{@{\extracolsep{\fill}}ccccccc@{}}
\toprule
\multirow{2}{*}{Caption} & \multicolumn{6}{c}{Evaluation Metric}\\
& CIDEr & CLIP-S & RefCLIP-S& CLAIR & ALOHa & Ours\\
\midrule[.05em]
Clean &  6.4 & 81.3 & 75.5 & \textbf{86.9} & 36.2& \textbf{62.8}\\
Object & 4.8 & 81.0 & 75.3 & 85.2 & 31.5& 52.3\\
Attribution & 6.2 & 80.9 & 75.2 & 80.0 & 34.3& 60.9\\
Relation & \textbf{6.7} & \textbf{81.4} & \textbf{75.6} & 83.5 & \textbf{36.9}& 51.9\\
\midrule[.1em]

\end{tabular*}
}
\label{tab:metaeval}
\end{table}
If a human were to measure the factuality of a text, it would be natural to decompose the text into units that can be classified as true or false, and then calculate the proportion of true units \citep{faithfulness}. We adopt this approach to measure the factuality of captions, utilizing the state-of-the-art model GPT-4o. In our framework, GPT-4o decomposes each caption into atomic propositions and determines their truthfulness based on the corresponding image and reference caption. If the number of atomic propositions judged as true and false are $T$ and $F$, respectively, the factuality of the caption is defined as $\frac{T}{T+F}$.

\begin{figure*}[t]
{
\begin{center}
\centerline{\includegraphics[width=1.0\textwidth]{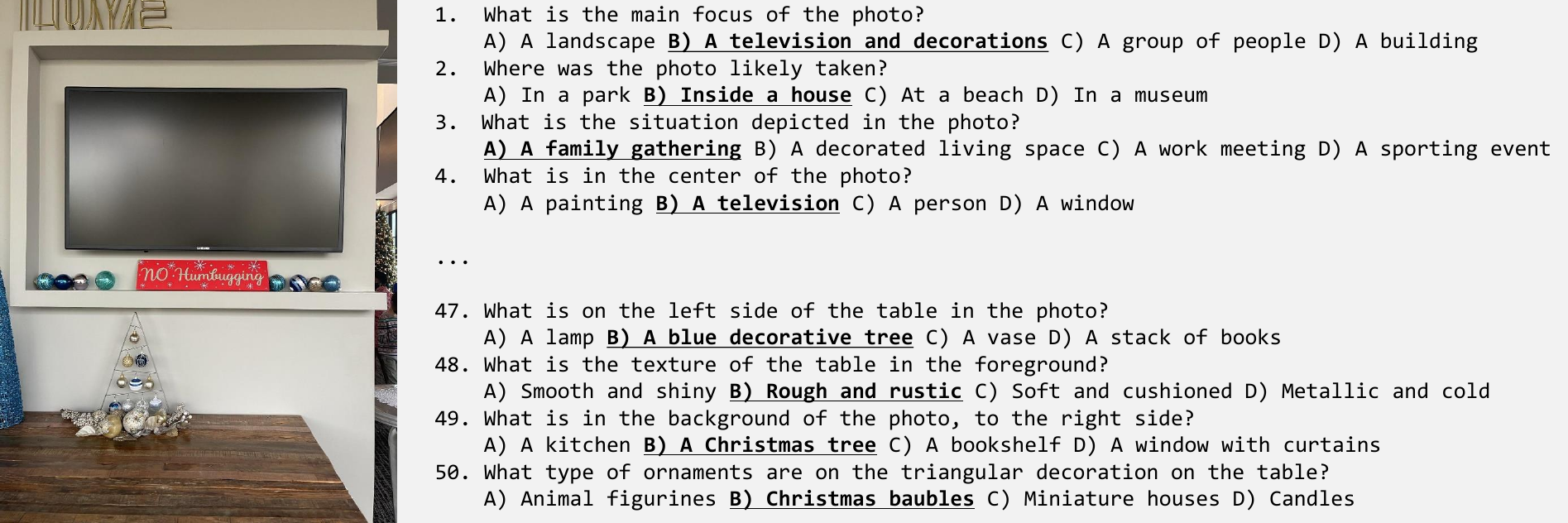}}%
\caption{
An example of our coverage evaluation data sample. The dataset consists of multiple-choice questions with four or fewer options. As demonstrated, the dataset includes questions with varying levels of granularity, ranging from broad to highly detailed. We have an LLM solve these problems using only the provided captions.
}
\label{fig:vqa}
\end{center}
}
\end{figure*}
To validate this evaluation method, we use the DOCCI dataset \citep{docci}, which contains human-annotated detailed image captions. Specifically, for each image in a subset of the dataset, we prepare the following four types of captions (details provided in Appendix \ref{app:prompts}):
\begin{enumerate}[leftmargin=*, nolistsep]
    \item Clean: The original caption (\eg, \emph{An indoor view captures a cat on a wooden floor, attempting to catch a large pale peacock feather flying above it}).
    \item Object: A description of an object likely present but not in the image is added to the Clean caption (\eg,  \emph{An indoor view captures a cat on a wooden floor, attempting to catch a large pale peacock feather flying above it. \textcolor{red}{A small red ball is rolling near the cat}}).
    \item Attribution: Some object attributions in the Clean caption are modified to be inconsistent with the image (\eg, \emph{An indoor view captures a cat on a \textcolor{red}{metal} floor, attempting to catch a \textcolor{red}{small dark} peacock feather flying above it}).
    \item Relation: Some object relationships in the Clean caption are altered to conflict with the image (\eg, \emph{An indoor view captures a cat on a wooden floor, attempting to catch a large pale peacock feather flying \textcolor{red}{below} it}).
\end{enumerate}

We evaluate the four types of captions using various metrics (BLEU, ROUGE, METEOR, CIDEr, CLIP-S, RefCLIP-S, CLAIR, and ALOHa), including our own, to determine whether the hallucinations in the three modified types are reflected in the scores. For a fair comparison, all methods requiring GPT (CLAIR, ALOHa, and ours) use GPT-4o, and all methods requiring reference captions (BLEU, ROUGE, METEOR, CIDEr, CLAIR, ALOHa, and ours) use a separate set \citep{iiw} of human-annotated captions.

Table \ref{tab:metaeval} shows that existing metrics are unreliable for evaluating the factuality of detailed image captions.
Specifically,
CLIP can only process up to 77 tokens and operates like a bag-of-words model \citep{bagofwords}. This prevents CLIP-based metrics from capturing the full content of detailed image captions, particularly missing Relation hallucinations. ALOHa effectively addresses Object and Attribution hallucinations but fails to capture Relation hallucinations due to its algorithmic limitations. CLAIR detects and reflects all three types of hallucinations in the scores. However, CLAIR does not focus solely on factuality; instead, it allows the GPT model to directly score each caption, applying the evaluation criteria implicitly defined by the GPT model. In contrast, our metric exclusively considers the factuality of the caption. While it does not assign a perfect score to the Clean captions due to GPT-4o's limitations in image understanding, it successfully assigns the highest score to Clean among the four caption sets.

\paragraph{Coverage.}
An image caption with only factual information is not high-quality if it focuses solely on trivial aspects of the image. To assess the coverage of captioning models, we propose a QA-based metric and a benchmark dataset. Our coverage evaluation method is based on the assumption that \emph{if an image caption fully captures the information in the image, visual questions about that image should be answerable by referencing the caption alone}.

Our goal is to evaluate hyper-detailed image captions. Therefore, the visual questions for evaluation must include a variety of detailed and nuanced questions about the images. Given the limitations of existing VQA datasets in this regard \citep{mme,pope,mmmu}, we construct a new VQA dataset.
However, creating a new VQA dataset that includes a variety of detailed questions requires substantial labor. To reduce the associated costs, we follow the process outlined below to construct our dataset:
\begin{enumerate}[leftmargin=*, nolistsep]
    \item Generating more than 50 questions per image in the IIW-400 dataset using GPT-4o.
    \item Deduplicating the questions for each image using Sentence-BERT \citep{sentence-bert}.
    \item Human labelers refine or remove ambiguous, flawed, or common-knowledge questions.
    \item Human labelers annotate the correct answers to the remaining and revised questions.
\end{enumerate}
Our coverage evaluation dataset contains a total of 19,899 multiple-choice questions, with each image averaging 49.8 questions.
We present an example of our dataset in Figure \ref{fig:vqa}. While our benchmark dataset can also be used to assess the visual understanding capabilities of MLLMs, we use it to evaluate the coverage of captioning models by having an LLM answer the questions based on the generated captions.
\section{Experimental Results and Discussion}
\subsection{Experimental Setup}
We adopt LLaVA-v1.5-7B, LLaVA-NeXT-7B, LLaVA-NeXT-13B, InternVL-Chat-V1.5, and GPT-4V as the models for both captioning and CapMAS’s fact-checking. We use LLaMA-3-8B \citep{llama3} or GPT-4 as the decomposer and corrector LLMs in CapMAS. Our experiments utilize the IIW-400 and DOCCI datasets, which contain images paired with highly detailed, hallucination-free captions. These high-quality reference captions enable precise evaluation of the captioning models.

We employ our proposed factuality and coverage metrics, along with CLAIR, all based on GPT-4o, to evaluate the generated captions. To ensure robust evaluation, we summarize the captions \citep{vfc} generated from five different prompts using LLaMA-3-8B. The only CapMAS hyperparameter, $\pi$, is tuned on a validation set of five examples sampled from the DCI dataset \citep{dci}. The prompt templates are provided in Appendix \ref{app:prompts}.
\subsection{Our Metric's Correlation with Human Evaluation}\label{sec:human_eval}
We obtain human evaluation data to validate the reliability of our factuality metric and compare it with existing ones. Human labelers assess which caption---LLaVA-v1.5-7B’s or InstructBLIP’s---is more factual for each image in a subset of the DOCCI test set (refer to Appendix \ref{app:human_eval_data} for further details). Using this dataset, we compare our metric with FAITHSCORE \citep{faithscore} and FACTSCORE \citep{factscore}, both of which evaluate the factuality of decomposed units: FactScore uses only the reference caption, FaithScore only the image, and our metric combines both.

Table \ref{tab:human_eval} shows that our metric exhibits a stronger correlation with human evaluation than existing metrics. However, this is not the sole reason to adopt our proposed metric for evaluating the factuality of detailed image captions. Metrics that rely solely on unimodal information are inherently susceptible to undesirable biases. For instance, metrics like FACTSCORE, which depend exclusively on reference captions, introduce stylistic biases tied to the specific style, tone, or phrasing of the references, unfairly favoring or penalizing captions based on these factors. In contrast, as demonstrated in Appendix \ref{app:human_eval_critical}, our metric is free from such biases.\begin{table}[]
\centering
\caption{Comparison of correlations between human preferences and automated metrics in terms of factuality.}\label{tab:human_eval}
{
\begin{tabular}{c @{\extracolsep{\fill}} ccc}
& FAITHSCORE & FACTSCORE & Ours\\\midrule[.05em]
Spearman's $\rho$& 62.5 & 67.9 & \textbf{70.2}\\
\midrule[.1em]
\end{tabular}
}
\end{table}
\subsection{Improving Captioning Model Factuality}\definecolor{LightCyan}{rgb}{0.88,1,1}
\newcolumntype{a}{>{\columncolor{LightCyan}}c}
\begin{table*}[]
\centering
    \caption{Effectiveness of our proposed method across various captioning models. In the CapMAS column, the LLM represents the decomposer and corrector, while the MLLM represents the fact-checker. Avg. denotes the average of CLAIR, Factuality, and Coverage.}
{
\begin{tabular*}{\textwidth}{@{\extracolsep{\fill}}cccccca@{}}
\toprule
\multirow{2}{*}{Captioner} & \multicolumn{2}{c}{CapMAS} & \multicolumn{4}{c}{Metric}\\
& LLM & MLLM & CLAIR & Factuality & Coverage & Avg.\\\midrule[.05em]
\multirow{3}{*}{LLaVA-NeXT-7B} & - & - & 68.8 &	59.9 &	\textbf{47.9} &	58.9\\
& LLaMA-3-8B & LLaVA-NeXT-7B & 74.1 & 72.2 & 46.9 & 64.4\\
& GPT-4 & LLaVA-NeXT-7B & \textbf{74.6} & \textbf{73.4} & 46.2 & \textbf{64.7}\\\midrule[.05em]
\multirow{3}{*}{LLaVA-NeXT-13B} & - & - & 70.2 & 62.1 & \textbf{48.5} & 60.3\\
& LLaMA-3-8B & LLaVA-NeXT-13B & \textbf{75.5} & 77.9 & 45.8 & \textbf{66.4}\\
& GPT-4 & LLaVA-NeXT-13B & 73.4 & \textbf{79.3} & 45.1 & 65.9\\\midrule[.05em]
\multirow{3}{*}{InternVL-Chat-V1.5} & - & - & 74.9 & 65.5 & \textbf{48.2} & 62.9\\
& LLaMA-3-8B & InternVL-Chat-V1.5 & \textbf{78.2} & \textbf{75.9} & 47.3 & \textbf{67.1}\\
& GPT-4 & InternVL-Chat-V1.5 & 77.8 & 75.7 & 47.3 & 66.9\\\midrule[.05em]
\multirow{4}{*}{GPT-4V} & - & - & 82.4 & 77.1 & \textbf{53.5} & 71.0\\
& LLaMA-3-8B & LLaVA-NeXT-7B & 83.3 & 83.3 & 50.8 & 72.4\\
& LLaMA-3-8B & LLaVA-NeXT-13B & 81.9 & \textbf{85.3} & 48.4 & 71.9\\
& LLaMA-3-8B & InternVL-Chat-V1.5 & \textbf{84.6} & 82.1 & \textbf{53.5} & \textbf{73.4}\\
\bottomrule

\end{tabular*}
}
\label{tab:facter}
\end{table*}
Our proposed CapMAS exhibits a loose factuality-coverage trade-off depending on the hyperparameter $\pi$. Specifically, as $\pi$ decreases, the threshold for determining factual propositions becomes stricter, leading to more propositions being identified for correction. Consequently, factuality increases while coverage decreases (an ablation study on $\pi$ is provided in Appendix \ref{app:ablation_study}). We first investigate whether CapMAS can enhance the factuality of various MLLMs while minimizing the reduction in coverage.

Table \ref{tab:facter} demonstrates that \textbf{CapMAS can significantly enhance the factuality of all tested MLLMs while minimizing coverage loss}. The substantial improvement in factuality, compared to the relatively minor coverage loss in the captioning models, is also reflected in the increased CLAIR scores.
\textbf{Using a more advanced LLM in CapMAS does not necessarily result in greater performance gains}. When applying CapMAS to the LLaVA and InternVL models, there is minimal difference between the results obtained with LLaMA-3-8B and those with GPT-4. This suggests that the LLM's role in CapMAS is relatively straightforward.
\textbf{CapMAS can improve detailed image captioning even for the state-of-the-art MLLM, GPT-4V}. It can significantly enhance factuality even when used with MLLMs far less capable than GPT-4V.
However, in such cases, there is a considerable loss in coverage, as many visual elements recognized by GPT-4V are identified as hallucinations by CapMAS. With InternVL-Chat-V1.5, CapMAS maintains GPT-4V's coverage while improving factuality.
We additionally provide a qualitative comparison in Figure \ref{fig:case} between LLaVA-NeXT-7B with and without the application of CapMAS (referencing the first two rows of Table \ref{tab:facter}).
\subsection{Comparison with Other Methods}\begin{table}[]
\centering
    \caption{Performance comparison between our proposed method and other methods regarding detailed image captioning. Base refers to the default image captioning of LLaVA-v1.5-7B.}
{
\footnotesize
\addtolength{\tabcolsep}{-4.3pt}
\begin{tabular}{cccca}
\toprule
Method & CLAIR & Factuality & Coverage & Avg.\\\midrule[.05em]
Base & 62.1 & 52.8 & 34.3 & 49.7 \\ 
VCD \citep{vcd} & 59.7 & 44.6 & 39.3 & 47.9 \\
OPERA \citep{opera} & 59.1 & 53.0 & 34.1 & 48.7\\
LURE \citep{lure} & 57.2 & 51.9 & 27.6 & 45.6\\
Volcano \citep{volcano} & 63.9 & 53.7 & 37.7 & 51.7 \\
LRV \citep{robust_instruction_tuning} & 39.7 & 29.1 & 37.8 & 35.5 \\
SPARC \citep{sparc} & 64.7 & 50.2 & \textbf{44.9} & 53.3 \\
CapMAS (ours) & \textbf{66.3} & \textbf{63.4} & 33.1 & \textbf{54.3}\\
\bottomrule

\end{tabular}
}
\label{tab:comparison}
\end{table}
Various methods have been proposed to mitigate hallucinations in MLLMs, and they have primarily been validated on VQA and simple captioning benchmarks. We compare CapMAS with three recent decoding-based methods (VCD, OPERA, and SPARC \citep{sparc}), two corrector-based methods (LURE and Volcano), and one training-based method (LRV) from the perspective of detailed image captioning. All methods, except for LRV and LURE, use LLaVA-v1.5-7B, while the LRV and LURE methods employ the MiniGPT-4 model \citep{minigpt4} as provided by their respective authors.
For reference, VisualFactChecker (VFC) \citep{vfc} is also a pipeline composed of pre-trained models that revise initial captions, similar to our approach. However, the inability to reproduce VFC, as its authors have not provided the necessary resources for reproduction, prevents a direct comparison with our method. Nonetheless, we can infer that our method outperforms VFC because 1) VFC specifically targets object hallucinations, and 2) it employs an object detector \citep{groundingdino} for hallucination detection (see Table \ref{tab:motivation}).

Table \ref{tab:comparison} shows that the decoding-based methods are ineffective for detailed image captioning. Ironically, applying VCD significantly reduces the factuality of the LLaVA model while increasing coverage. SPARC does not show a notable impact in terms of factuality, but it proves to be considerably beneficial with respect to coverage. Volcano yields only slight improvements in LLaVA’s captions. However, CapMAS substantially enhances the factuality of the captioning model compared to the other methods. \emph{These results suggest that methods proposed to enhance MLLM factuality should be evaluated not only on tasks requiring short responses, such as VQA, but also on detailed image captioning tasks}.
\subsection{Consistency Between MLLM Captioning and VQA Evaluation results}\begin{table*}[]
\centering
    \caption{Detailed image captioning and VQA performance of various MLLMs. OpenCompass \citep{opencompass} includes MMBench v1.1 \citep{mmbench}, MMStar \citep{mmstar}, MMMU val \citep{mmmu}, MathVista \citep{mathvista}, OCRBench \citep{ocrbench}, AI2D \citep{ai2d}, HallusionBench \citep{halbench}, and MMVet \citep{mmvet}. For POPE \citep{pope}, we report the average F1 score across the three categories: adversarial, popular, and random. We report the sum of the perception and cognition scores for MME \citep{mme}. The best results for each metric are shown in \textbf{bold}.}
{
\addtolength{\tabcolsep}{-3.5pt}
\begin{tabular*}{\textwidth}{@{\extracolsep{\fill}}ccccaccca@{}}
\toprule
\multirow{2}{*}{Model} & \multicolumn{4}{c}{Detailed Image Captioning} & \multicolumn{4}{c}{Visual Question Answering}\\
\cmidrule(lr){2-5}\cmidrule(lr){6-9}
& CLAIR & Factuality & Coverage & Avg. & OpenCompass & MME & POPE & Avg.\\\midrule[.05em]
InstructBLIP-7B & 57.2 &	44.4 &	30.3 &	43.9 &	31.1 &	1391.4 &	86.1 &	38.4\\
LLaVA-v1.5-7B & 61.1 &	56.3 &	30.5 &	49.3 &	36.9 &	1808.4 &	86.1 &	44.6\\
LLaVA-NeXT-7B & 63.8 &	58.5 &	42.2 &	54.8 &	44.7 &	1769.1 &	87.5 &	50.8\\
LLaVA-NeXT-13B & 64.5 &	62.8 &	43.0 &	56.8 &	47.6 &	1745.6 &	\textbf{87.8} &	53.1\\
Idefics2-8B & 58.1 & \textbf{85.2} & 13.4 &	52.2 &	53.0 &	1847.6 &	86.2 &	57.6\\
InternVL-Chat-V1.5 & 72.4 &	67.6 &	46.0 &	62.0 &	61.7 &	2189.6 &	87.5 &	65.9\\
MiniCPM-V-2.6 & 73.1 &	68.9 &	43.6 &	61.9 &	\textbf{65.2} &	\textbf{2268.7} &	83.2 &	\textbf{68.6}\\
GPT-4V & \textbf{82.4} & 78.6 &	\textbf{52.6} &	\textbf{71.2} &	63.5 &	2070.2 &	81.8 &	66.4\\

\midrule[.1em]

\end{tabular*}
}
\label{tab:vqa_captioning}
\end{table*}
Currently, MLLM evaluations are conducted on tasks that require only short responses, such as VQA tasks \citep{opencompass}. However, to assess the potential of MLLMs in real-world applications, such as visual assistants, it is essential to evaluate their detailed image captioning abilities. The ranking of models used in our experiments, including LLaVA-v1.5-7B, LLaVA-NeXT-7B, LLaVA-NeXT-13B, InternVL-Chat-V1.5, and GPT-4V, is consistent across both our captioning evaluation results and widely used benchmarks like MMMU \citep{mmmu}.
However, for instance, some MLLMs may be optimized for VQA tasks that require only short responses, allowing them to rank highly on common VQA benchmarks, yet their limited image captioning abilities could restrict their practical use. To investigate this, we evaluate the detailed image captioning capabilities of various MLLMs and examine whether their rankings are consistent with their rankings on widely used VQA benchmarks. We adopt InstructBLIP-7B \citep{instructblip}, Idefics2-8B \citep{idefics2}, and MiniCPM-V-2.6 \citep{minicpm} as additional MLLMs for the experiment.

Table \ref{tab:vqa_captioning} presents the evaluation results of MLLMs' responses to the prompt ``Describe the given image in a very detailed manner'' as well as the performance of these models on various VQA tasks. From these results, we observe that the performance of an MLLM on widely used benchmarks does not necessarily reflect its capabilities in detailed image captioning. Specifically, Idefics2-8B ranks mid-tier among the tested models in VQA tasks but falls into the lowest-performing group in terms of detailed image captioning. Its high factuality but low coverage indicates that Idefics2-8B has been trained to provide short and concise answers; this conclusion remains unchanged even when using Idefics2-8B-Chatty \citep{idefics2}. Despite being a relatively small model, MiniCPM-V-2.6 attracted attention by outperforming GPT-4V on benchmarks. However, our results show that the model significantly underperforms GPT-4V in detailed image captioning. Additionally, we find that the factuality of the captions cannot be reliably predicted from the accuracy of MLLMs on POPE \citep{pope}, which was proposed to evaluate object hallucinations.

\emph{Based on these experimental results, we raise concerns about the current MLLM evaluations that are centered around VQA tasks. We encourage the community to also evaluate MLLMs from the perspective of detailed image captioning in order to showcase their full potential.}

We provide additional results on fluency, cost, and test-time scaling in Appendix \ref{app:addexp}.
\section{Conclusion}
Detailed image captioning tasks are closely linked to critical applications, such as visual assistance for the impaired. Our research aims to assess and enhance the potential of MLLMs in these real-world contexts. We propose CapMAS, a method that improves detailed image captions through the collaboration of a pre-trained MLLM and LLM. In addition, we introduce a framework and benchmark dataset for evaluating the factuality and coverage of captioning models. Our experiments validate the proposed evaluation framework and demonstrate that CapMAS significantly improves the factuality of captioning models.
We additionally present the following two key observations:
\begin{itemize}[leftmargin=*, nolistsep]
\item Methods designed to improve MLLM factuality, which have been validated primarily on VQA or short captioning tasks, may be ineffective for detailed image captioning and can even reduce the factuality of the backbone model.
\item High performance on commonly used VQA-centric benchmarks does not necessarily indicate that the model will excel in hyper-detailed image captioning.
\end{itemize}
These observations raise concerns about the current VQA-centric trend in MLLM evaluation. We encourage the community to evaluate MLLMs and related algorithms not only on VQA tasks but also on detailed image captioning tasks to gain a more comprehensive understanding of their potential.

\section*{Impact Statement}

This research contributes to developing more accurate and reliable image captioning systems, which are crucial for accessibility technologies. The proposed multiagent approach mitigates the risks of misinformation and hallucinations in AI-generated content, enhancing the safety and trustworthiness of AI systems. However, as image captioning models become more detailed, ethical concerns, particularly privacy-related, may emerge. Future research should address these challenges to ensure the responsible deployment of this technology.

\section*{Acknowledgements}

This work was supported by Institute of Information \& communications Technology Planning \& Evaluation (IITP) grant funded by the Korea government (MSIT) [No.RS-2021-II211343, RS-2022-II220959, Artificial Intelligence Graduate School Program (Seoul National University), No.RS-2025-02263754, Human-Centric Embodied AI Agents with Autonomous Decision-Making], the National Research Foundation of Korea (NRF) grant funded by the Korea government (MSIT) (No. 2022R1A3B1077720), a grant from Yang Young Foundation, and the BK21 FOUR program of the Education and the Research Program for Future ICT Pioneers, Seoul National University in 2025.

\bibliography{example_paper}

\begin{thebibliography}{60}
\providecommand{\natexlab}[1]{#1}
\providecommand{\url}[1]{\texttt{#1}}
\expandafter\ifx\csname urlstyle\endcsname\relax
  \providecommand{\doi}[1]{doi: #1}\else
  \providecommand{\doi}{doi: \begingroup \urlstyle{rm}\Url}\fi

\bibitem[Achiam et~al.(2023)Achiam, Adler, Agarwal, Ahmad, Akkaya, Aleman, Almeida, Altenschmidt, Altman, Anadkat, et~al.]{gpt4}
Achiam, J., Adler, S., Agarwal, S., Ahmad, L., Akkaya, I., Aleman, F.~L., Almeida, D., Altenschmidt, J., Altman, S., Anadkat, S., et~al.
\newblock Gpt-4 technical report.
\newblock \emph{arXiv preprint arXiv:2303.08774}, 2023.

\bibitem[AI@Meta(2024)]{llama3}
AI@Meta.
\newblock Llama 3 model card.
\newblock 2024.
\newblock URL \url{https://github.com/meta-llama/llama3/blob/main/MODEL_CARD.md}.

\bibitem[Banerjee \& Lavie(2005)Banerjee and Lavie]{meteor}
Banerjee, S. and Lavie, A.
\newblock Meteor: An automatic metric for mt evaluation with improved correlation with human judgments.
\newblock In \emph{Proceedings of the acl workshop on intrinsic and extrinsic evaluation measures for machine translation and/or summarization}, pp.\  65--72, 2005.

\bibitem[Brown et~al.(2020)Brown, Mann, Ryder, Subbiah, Kaplan, Dhariwal, Neelakantan, Shyam, Sastry, Askell, et~al.]{gpt3}
Brown, T.~B., Mann, B., Ryder, N., Subbiah, M., Kaplan, J., Dhariwal, P., Neelakantan, A., Shyam, P., Sastry, G., Askell, A., et~al.
\newblock Language models are few-shot learners.
\newblock \emph{arXiv preprint arXiv:2005.14165}, 2020.

\bibitem[Chan et~al.(2023)Chan, Petryk, Gonzalez, Darrell, and Canny]{clair}
Chan, D., Petryk, S., Gonzalez, J., Darrell, T., and Canny, J.
\newblock Clair: Evaluating image captions with large language models.
\newblock In \emph{Proceedings of the 2023 Conference on Empirical Methods in Natural Language Processing}, pp.\  13638--13646, 2023.

\bibitem[Chen et~al.(2024{\natexlab{a}})Chen, Li, Dong, Zhang, Zang, Chen, Duan, Wang, Qiao, Lin, et~al.]{mmstar}
Chen, L., Li, J., Dong, X., Zhang, P., Zang, Y., Chen, Z., Duan, H., Wang, J., Qiao, Y., Lin, D., et~al.
\newblock Are we on the right way for evaluating large vision-language models?
\newblock \emph{arXiv preprint arXiv:2403.20330}, 2024{\natexlab{a}}.

\bibitem[Chen et~al.(2024{\natexlab{b}})Chen, Wang, Tian, Ye, Gao, Cui, Tong, Hu, Luo, Ma, et~al.]{internvl_1_5}
Chen, Z., Wang, W., Tian, H., Ye, S., Gao, Z., Cui, E., Tong, W., Hu, K., Luo, J., Ma, Z., et~al.
\newblock How far are we to gpt-4v? closing the gap to commercial multimodal models with open-source suites.
\newblock \emph{arXiv preprint arXiv:2404.16821}, 2024{\natexlab{b}}.

\bibitem[Chen et~al.(2024{\natexlab{c}})Chen, Wu, Wang, Su, Chen, Xing, Zhong, Zhang, Zhu, Lu, et~al.]{internvl}
Chen, Z., Wu, J., Wang, W., Su, W., Chen, G., Xing, S., Zhong, M., Zhang, Q., Zhu, X., Lu, L., et~al.
\newblock Internvl: Scaling up vision foundation models and aligning for generic visual-linguistic tasks.
\newblock In \emph{Proceedings of the IEEE/CVF Conference on Computer Vision and Pattern Recognition}, pp.\  24185--24198, 2024{\natexlab{c}}.

\bibitem[Cui et~al.(2023)Cui, Yang, and Yao]{qllama}
Cui, Y., Yang, Z., and Yao, X.
\newblock Efficient and effective text encoding for chinese llama and alpaca.
\newblock \emph{arXiv preprint arXiv:2304.08177}, 2023.

\bibitem[Dai et~al.(2023)Dai, Li, Li, Tiong, Zhao, Wang, Li, Fung, and Hoi]{instructblip}
Dai, W., Li, J., Li, D., Tiong, A., Zhao, J., Wang, W., Li, B., Fung, P., and Hoi, S.
\newblock Instruct{BLIP}: Towards general-purpose vision-language models with instruction tuning.
\newblock In \emph{Thirty-seventh Conference on Neural Information Processing Systems}, 2023.
\newblock URL \url{https://openreview.net/forum?id=vvoWPYqZJA}.

\bibitem[Duan et~al.(2024)Duan, Yang, Qiao, Fang, Chen, Liu, Dong, Zang, Zhang, Wang, Lin, and Chen]{opencompass}
Duan, H., Yang, J., Qiao, Y., Fang, X., Chen, L., Liu, Y., Dong, X., Zang, Y., Zhang, P., Wang, J., Lin, D., and Chen, K.
\newblock Vlmevalkit: An open-source toolkit for evaluating large multi-modality models, 2024.
\newblock URL \url{https://arxiv.org/abs/2407.11691}.

\bibitem[Garg et~al.(2024)Garg, Burns, Ayan, Bitton, Montgomery, Onoe, Bunner, Krishna, Baldridge, and Soricut]{iiw}
Garg, R., Burns, A., Ayan, B.~K., Bitton, Y., Montgomery, C., Onoe, Y., Bunner, A., Krishna, R., Baldridge, J., and Soricut, R.
\newblock Imageinwords: Unlocking hyper-detailed image descriptions.
\newblock \emph{arXiv preprint arXiv:2405.02793}, 2024.

\bibitem[Ge et~al.(2024)Ge, Zeng, Huffman, Lin, Liu, and Cui]{vfc}
Ge, Y., Zeng, X., Huffman, J.~S., Lin, T.-Y., Liu, M.-Y., and Cui, Y.
\newblock Visual fact checker: Enabling high-fidelity detailed caption generation.
\newblock In \emph{Proceedings of the IEEE/CVF Conference on Computer Vision and Pattern Recognition}, pp.\  14033--14042, 2024.

\bibitem[Guan et~al.(2024)Guan, Liu, Wu, Xian, Li, Liu, Wang, Chen, Huang, Yacoob, et~al.]{halbench}
Guan, T., Liu, F., Wu, X., Xian, R., Li, Z., Liu, X., Wang, X., Chen, L., Huang, F., Yacoob, Y., et~al.
\newblock Hallusionbench: an advanced diagnostic suite for entangled language hallucination and visual illusion in large vision-language models.
\newblock In \emph{Proceedings of the IEEE/CVF Conference on Computer Vision and Pattern Recognition}, pp.\  14375--14385, 2024.

\bibitem[Hessel et~al.(2021)Hessel, Holtzman, Forbes, Le~Bras, and Choi]{clipscore}
Hessel, J., Holtzman, A., Forbes, M., Le~Bras, R., and Choi, Y.
\newblock Clipscore: A reference-free evaluation metric for image captioning.
\newblock In \emph{Proceedings of the 2021 Conference on Empirical Methods in Natural Language Processing}, pp.\  7514--7528, 2021.

\bibitem[Hossain et~al.(2019)Hossain, Sohel, Shiratuddin, and Laga]{caption_survey}
Hossain, M.~Z., Sohel, F., Shiratuddin, M.~F., and Laga, H.
\newblock A comprehensive survey of deep learning for image captioning.
\newblock \emph{ACM Computing Surveys (CsUR)}, 51\penalty0 (6):\penalty0 1--36, 2019.

\bibitem[Huang et~al.(2024)Huang, Dong, Zhang, Wang, He, Wang, Lin, Zhang, and Yu]{opera}
Huang, Q., Dong, X., Zhang, P., Wang, B., He, C., Wang, J., Lin, D., Zhang, W., and Yu, N.
\newblock Opera: Alleviating hallucination in multi-modal large language models via over-trust penalty and retrospection-allocation.
\newblock In \emph{Proceedings of the IEEE/CVF Conference on Computer Vision and Pattern Recognition}, pp.\  13418--13427, 2024.

\bibitem[Jing et~al.(2024)Jing, Li, Chen, and Du]{faithscore}
Jing, L., Li, R., Chen, Y., and Du, X.
\newblock {F}aith{S}core: Fine-grained evaluations of hallucinations in large vision-language models.
\newblock In Al-Onaizan, Y., Bansal, M., and Chen, Y.-N. (eds.), \emph{Findings of the Association for Computational Linguistics: EMNLP 2024}, pp.\  5042--5063, Miami, Florida, USA, November 2024. Association for Computational Linguistics.
\newblock \doi{10.18653/v1/2024.findings-emnlp.290}.
\newblock URL \url{https://aclanthology.org/2024.findings-emnlp.290}.

\bibitem[Jung et~al.(2025)Jung, Lee, Kim, and Yoon]{sparc}
Jung, M., Lee, S., Kim, E., and Yoon, S.
\newblock Visual attention never fades: Selective progressive attention recalibration for detailed image captioning in multimodal large language models.
\newblock \emph{arXiv preprint arXiv:2502.01419}, 2025.

\bibitem[Kembhavi et~al.(2016)Kembhavi, Salvato, Kolve, Seo, Hajishirzi, and Farhadi]{ai2d}
Kembhavi, A., Salvato, M., Kolve, E., Seo, M., Hajishirzi, H., and Farhadi, A.
\newblock A diagram is worth a dozen images.
\newblock In \emph{Computer Vision--ECCV 2016: 14th European Conference, Amsterdam, The Netherlands, October 11--14, 2016, Proceedings, Part IV 14}, pp.\  235--251. Springer, 2016.

\bibitem[Lauren{\c{c}}on et~al.(2024)Lauren{\c{c}}on, Tronchon, Cord, and Sanh]{idefics2}
Lauren{\c{c}}on, H., Tronchon, L., Cord, M., and Sanh, V.
\newblock What matters when building vision-language models?
\newblock \emph{arXiv preprint arXiv:2405.02246}, 2024.

\bibitem[Lee et~al.(2024)Lee, Park, Jo, and Seo]{volcano}
Lee, S., Park, S., Jo, Y., and Seo, M.
\newblock Volcano: Mitigating multimodal hallucination through self-feedback guided revision.
\newblock In \emph{Proceedings of the 2024 Conference of the North American Chapter of the Association for Computational Linguistics: Human Language Technologies (Volume 1: Long Papers)}, pp.\  391--404, 2024.

\bibitem[Leng et~al.(2024)Leng, Zhang, Chen, Li, Lu, Miao, and Bing]{vcd}
Leng, S., Zhang, H., Chen, G., Li, X., Lu, S., Miao, C., and Bing, L.
\newblock Mitigating object hallucinations in large vision-language models through visual contrastive decoding.
\newblock In \emph{Proceedings of the IEEE/CVF Conference on Computer Vision and Pattern Recognition}, pp.\  13872--13882, 2024.

\bibitem[Li et~al.(2023{\natexlab{a}})Li, Li, Savarese, and Hoi]{blip2}
Li, J., Li, D., Savarese, S., and Hoi, S.
\newblock Blip-2: Bootstrapping language-image pre-training with frozen image encoders and large language models.
\newblock In \emph{International conference on machine learning}, pp.\  19730--19742. PMLR, 2023{\natexlab{a}}.

\bibitem[Li et~al.(2023{\natexlab{b}})Li, Du, Zhou, Wang, Zhao, and Wen]{pope}
Li, Y., Du, Y., Zhou, K., Wang, J., Zhao, W.~X., and Wen, J.-R.
\newblock Evaluating object hallucination in large vision-language models.
\newblock \emph{arXiv preprint arXiv:2305.10355}, 2023{\natexlab{b}}.

\bibitem[Lin(2004)]{rouge}
Lin, C.-Y.
\newblock Rouge: A package for automatic evaluation of summaries.
\newblock In \emph{Text summarization branches out}, pp.\  74--81, 2004.

\bibitem[Liu et~al.(2023{\natexlab{a}})Liu, Lin, Li, Wang, Yacoob, and Wang]{robust_instruction_tuning}
Liu, F., Lin, K., Li, L., Wang, J., Yacoob, Y., and Wang, L.
\newblock Mitigating hallucination in large multi-modal models via robust instruction tuning.
\newblock In \emph{The Twelfth International Conference on Learning Representations}, 2023{\natexlab{a}}.

\bibitem[Liu et~al.(2024{\natexlab{a}})Liu, Li, Li, Li, Zhang, Shen, and Lee]{llavanext}
Liu, H., Li, C., Li, Y., Li, B., Zhang, Y., Shen, S., and Lee, Y.~J.
\newblock Llava-next: Improved reasoning, ocr, and world knowledge, January 2024{\natexlab{a}}.
\newblock URL \url{https://llava-vl.github.io/blog/2024-01-30-llava-next/}.

\bibitem[Liu et~al.(2024{\natexlab{b}})Liu, Li, Wu, and Lee]{llava}
Liu, H., Li, C., Wu, Q., and Lee, Y.~J.
\newblock Visual instruction tuning.
\newblock \emph{Advances in neural information processing systems}, 36, 2024{\natexlab{b}}.

\bibitem[Liu et~al.(2023{\natexlab{b}})Liu, Zeng, Ren, Li, Zhang, Yang, Li, Yang, Su, Zhu, et~al.]{groundingdino}
Liu, S., Zeng, Z., Ren, T., Li, F., Zhang, H., Yang, J., Li, C., Yang, J., Su, H., Zhu, J., et~al.
\newblock Grounding dino: Marrying dino with grounded pre-training for open-set object detection.
\newblock \emph{arXiv preprint arXiv:2303.05499}, 2023{\natexlab{b}}.

\bibitem[Liu et~al.(2024{\natexlab{c}})Liu, Zheng, and Chen]{hypothesis1}
Liu, S., Zheng, K., and Chen, W.
\newblock Paying more attention to image: A training-free method for alleviating hallucination in lvlms.
\newblock \emph{arXiv preprint arXiv:2407.21771}, 2024{\natexlab{c}}.

\bibitem[Liu et~al.(2023{\natexlab{c}})Liu, Duan, Zhang, Li, Zhang, Zhao, Yuan, Wang, He, Liu, et~al.]{mmbench}
Liu, Y., Duan, H., Zhang, Y., Li, B., Zhang, S., Zhao, W., Yuan, Y., Wang, J., He, C., Liu, Z., et~al.
\newblock Mmbench: Is your multi-modal model an all-around player?
\newblock \emph{arXiv preprint arXiv:2307.06281}, 2023{\natexlab{c}}.

\bibitem[Liu et~al.(2024{\natexlab{d}})Liu, Li, Huang, Yang, Yu, Li, Yin, lin Liu, Jin, and Bai]{ocrbench}
Liu, Y., Li, Z., Huang, M., Yang, B., Yu, W., Li, C., Yin, X., lin Liu, C., Jin, L., and Bai, X.
\newblock On the hidden mystery of ocr in large multimodal models, 2024{\natexlab{d}}.
\newblock URL \url{https://arxiv.org/abs/2305.07895}.

\bibitem[Lu et~al.(2024)Lu, Bansal, Xia, Liu, Li, Hajishirzi, Cheng, Chang, Galley, and Gao]{mathvista}
Lu, P., Bansal, H., Xia, T., Liu, J., Li, C., Hajishirzi, H., Cheng, H., Chang, K.-W., Galley, M., and Gao, J.
\newblock Mathvista: Evaluating mathematical reasoning of foundation models in visual contexts.
\newblock In \emph{International Conference on Learning Representations (ICLR)}, 2024.

\bibitem[Madaan et~al.(2023)Madaan, Tandon, Gupta, Hallinan, Gao, Wiegreffe, Alon, Dziri, Prabhumoye, Yang, et~al.]{self_refine}
Madaan, A., Tandon, N., Gupta, P., Hallinan, S., Gao, L., Wiegreffe, S., Alon, U., Dziri, N., Prabhumoye, S., Yang, Y., et~al.
\newblock Self-refine: Iterative refinement with self-feedback.
\newblock \emph{Advances in Neural Information Processing Systems}, 36:\penalty0 46534--46594, 2023.

\bibitem[Maynez et~al.(2020)Maynez, Narayan, Bohnet, and McDonald]{faithfulness}
Maynez, J., Narayan, S., Bohnet, B., and McDonald, R.
\newblock On faithfulness and factuality in abstractive summarization.
\newblock In \emph{Proceedings of the 58th Annual Meeting of the Association for Computational Linguistics}, pp.\  1906--1919, 2020.

\bibitem[Min et~al.(2023)Min, Krishna, Lyu, Lewis, Yih, Koh, Iyyer, Zettlemoyer, and Hajishirzi]{factscore}
Min, S., Krishna, K., Lyu, X., Lewis, M., Yih, W.-t., Koh, P., Iyyer, M., Zettlemoyer, L., and Hajishirzi, H.
\newblock {FA}ct{S}core: Fine-grained atomic evaluation of factual precision in long form text generation.
\newblock In Bouamor, H., Pino, J., and Bali, K. (eds.), \emph{Proceedings of the 2023 Conference on Empirical Methods in Natural Language Processing}, pp.\  12076--12100, Singapore, December 2023. Association for Computational Linguistics.
\newblock \doi{10.18653/v1/2023.emnlp-main.741}.
\newblock URL \url{https://aclanthology.org/2023.emnlp-main.741}.

\bibitem[Onoe et~al.(2024)Onoe, Rane, Berger, Bitton, Cho, Garg, Ku, Parekh, Pont-Tuset, Tanzer, et~al.]{docci}
Onoe, Y., Rane, S., Berger, Z., Bitton, Y., Cho, J., Garg, R., Ku, A., Parekh, Z., Pont-Tuset, J., Tanzer, G., et~al.
\newblock Docci: Descriptions of connected and contrasting images.
\newblock \emph{arXiv preprint arXiv:2404.19753}, 2024.

\bibitem[Papineni et~al.(2002)Papineni, Roukos, Ward, and Zhu]{bleu}
Papineni, K., Roukos, S., Ward, T., and Zhu, W.-J.
\newblock Bleu: a method for automatic evaluation of machine translation.
\newblock In \emph{Proceedings of the 40th annual meeting of the Association for Computational Linguistics}, pp.\  311--318, 2002.

\bibitem[Petryk et~al.(2024)Petryk, Chan, Kachinthaya, Zou, Canny, Gonzalez, and Darrell]{aloha}
Petryk, S., Chan, D., Kachinthaya, A., Zou, H., Canny, J., Gonzalez, J., and Darrell, T.
\newblock Aloha: A new measure for hallucination in captioning models.
\newblock In \emph{Proceedings of the 2024 Conference of the North American Chapter of the Association for Computational Linguistics: Human Language Technologies (Volume 2: Short Papers)}, pp.\  342--357, 2024.

\bibitem[Radford et~al.(2021)Radford, Kim, Hallacy, Ramesh, Goh, Agarwal, Sastry, Askell, Mishkin, Clark, et~al.]{clip}
Radford, A., Kim, J.~W., Hallacy, C., Ramesh, A., Goh, G., Agarwal, S., Sastry, G., Askell, A., Mishkin, P., Clark, J., et~al.
\newblock Learning transferable visual models from natural language supervision.
\newblock In \emph{International conference on machine learning}, pp.\  8748--8763. PMLR, 2021.

\bibitem[Reimers \& Gurevych(2019)Reimers and Gurevych]{sentence-bert}
Reimers, N. and Gurevych, I.
\newblock Sentence-bert: Sentence embeddings using siamese bert-networks.
\newblock In \emph{Proceedings of the 2019 Conference on Empirical Methods in Natural Language Processing}. Association for Computational Linguistics, 11 2019.
\newblock URL \url{https://arxiv.org/abs/1908.10084}.

\bibitem[Urbanek et~al.(2024)Urbanek, Bordes, Astolfi, Williamson, Sharma, and Romero-Soriano]{dci}
Urbanek, J., Bordes, F., Astolfi, P., Williamson, M., Sharma, V., and Romero-Soriano, A.
\newblock A picture is worth more than 77 text tokens: Evaluating clip-style models on dense captions.
\newblock In \emph{Proceedings of the IEEE/CVF Conference on Computer Vision and Pattern Recognition}, pp.\  26700--26709, 2024.

\bibitem[Vaswani(2017)]{transformer}
Vaswani, A.
\newblock Attention is all you need.
\newblock \emph{Advances in Neural Information Processing Systems}, 2017.

\bibitem[Vedantam et~al.(2015)Vedantam, Lawrence~Zitnick, and Parikh]{cider}
Vedantam, R., Lawrence~Zitnick, C., and Parikh, D.
\newblock Cider: Consensus-based image description evaluation.
\newblock In \emph{Proceedings of the IEEE conference on computer vision and pattern recognition}, pp.\  4566--4575, 2015.

\bibitem[Vinyals et~al.(2015)Vinyals, Toshev, Bengio, and Erhan]{show_and_tell}
Vinyals, O., Toshev, A., Bengio, S., and Erhan, D.
\newblock Show and tell: A neural image caption generator.
\newblock In \emph{Proceedings of the IEEE conference on computer vision and pattern recognition}, pp.\  3156--3164, 2015.

\bibitem[Wang et~al.(2023)Wang, Wei, Schuurmans, Le, Chi, Narang, Chowdhery, and Zhou]{self_consistency}
Wang, X., Wei, J., Schuurmans, D., Le, Q.~V., Chi, E.~H., Narang, S., Chowdhery, A., and Zhou, D.
\newblock Self-consistency improves chain of thought reasoning in language models.
\newblock In \emph{The Eleventh International Conference on Learning Representations}, 2023.

\bibitem[Xu et~al.(2015)Xu, Ba, Kiros, Cho, Courville, Salakhudinov, Zemel, and Bengio]{show_attend_and_tell}
Xu, K., Ba, J., Kiros, R., Cho, K., Courville, A., Salakhudinov, R., Zemel, R., and Bengio, Y.
\newblock Show, attend and tell: Neural image caption generation with visual attention.
\newblock In \emph{International conference on machine learning}, pp.\  2048--2057. PMLR, 2015.

\bibitem[Yao et~al.(2024)Yao, Yu, Zhang, Wang, Cui, Zhu, Cai, Li, Zhao, He, et~al.]{minicpm}
Yao, Y., Yu, T., Zhang, A., Wang, C., Cui, J., Zhu, H., Cai, T., Li, H., Zhao, W., He, Z., et~al.
\newblock Minicpm-v: A gpt-4v level mllm on your phone.
\newblock \emph{arXiv preprint arXiv:2408.01800}, 2024.

\bibitem[Yin et~al.(2023{\natexlab{a}})Yin, Fu, Zhao, Li, Sun, Xu, and Chen]{mllm_survey}
Yin, S., Fu, C., Zhao, S., Li, K., Sun, X., Xu, T., and Chen, E.
\newblock A survey on multimodal large language models.
\newblock \emph{arXiv preprint arXiv:2306.13549}, 2023{\natexlab{a}}.

\bibitem[Yin et~al.(2023{\natexlab{b}})Yin, Fu, Zhao, Li, Sun, Xu, and Chen]{mme}
Yin, S., Fu, C., Zhao, S., Li, K., Sun, X., Xu, T., and Chen, E.
\newblock A survey on multimodal large language models.
\newblock \emph{arXiv preprint arXiv:2306.13549}, 2023{\natexlab{b}}.

\bibitem[Yin et~al.(2023{\natexlab{c}})Yin, Fu, Zhao, Xu, Wang, Sui, Shen, Li, Sun, and Chen]{woodpecker}
Yin, S., Fu, C., Zhao, S., Xu, T., Wang, H., Sui, D., Shen, Y., Li, K., Sun, X., and Chen, E.
\newblock Woodpecker: Hallucination correction for multimodal large language models.
\newblock \emph{arXiv preprint arXiv:2310.16045}, 2023{\natexlab{c}}.

\bibitem[Yu et~al.(2023)Yu, Yang, Li, Wang, Lin, Liu, Wang, and Wang]{mmvet}
Yu, W., Yang, Z., Li, L., Wang, J., Lin, K., Liu, Z., Wang, X., and Wang, L.
\newblock Mm-vet: Evaluating large multimodal models for integrated capabilities.
\newblock \emph{arXiv preprint arXiv:2308.02490}, 2023.

\bibitem[Yue et~al.(2024)Yue, Ni, Zhang, Zheng, Liu, Zhang, Stevens, Jiang, Ren, Sun, Wei, Yu, Yuan, Sun, Yin, Zheng, Yang, Liu, Huang, Sun, Su, and Chen]{mmmu}
Yue, X., Ni, Y., Zhang, K., Zheng, T., Liu, R., Zhang, G., Stevens, S., Jiang, D., Ren, W., Sun, Y., Wei, C., Yu, B., Yuan, R., Sun, R., Yin, M., Zheng, B., Yang, Z., Liu, Y., Huang, W., Sun, H., Su, Y., and Chen, W.
\newblock Mmmu: A massive multi-discipline multimodal understanding and reasoning benchmark for expert agi.
\newblock In \emph{Proceedings of CVPR}, 2024.

\bibitem[Yuksekgonul et~al.(2023)Yuksekgonul, Bianchi, Kalluri, Jurafsky, and Zou]{bagofwords}
Yuksekgonul, M., Bianchi, F., Kalluri, P., Jurafsky, D., and Zou, J.
\newblock When and why vision-language models behave like bags-of-words, and what to do about it?
\newblock In \emph{The Eleventh International Conference on Learning Representations}, 2023.

\bibitem[Zhang et~al.(2023)Zhang, Qiu, Guo, Deng, Zhang, Zhang, Zhou, Wang, and Fu]{confidence1}
Zhang, T., Qiu, L., Guo, Q., Deng, C., Zhang, Y., Zhang, Z., Zhou, C., Wang, X., and Fu, L.
\newblock Enhancing uncertainty-based hallucination detection with stronger focus.
\newblock In \emph{Proceedings of the 2023 Conference on Empirical Methods in Natural Language Processing}, pp.\  915--932, 2023.

\bibitem[Zhao et~al.(2024)Zhao, Yan, Sun, Xing, Meng, Wang, Cheng, Ren, and Yin]{zhao2024knowing}
Zhao, Y., Yan, L., Sun, W., Xing, G., Meng, C., Wang, S., Cheng, Z., Ren, Z., and Yin, D.
\newblock Knowing what llms do not know: A simple yet effective self-detection method.
\newblock In \emph{Proceedings of the 2024 Conference of the North American Chapter of the Association for Computational Linguistics: Human Language Technologies (Volume 1: Long Papers)}, pp.\  7044--7056, 2024.

\bibitem[Zhong et~al.(2024)Zhong, Feng, Zhao, Li, Huang, Gu, Ma, Xu, and Qin]{hypothesis2}
Zhong, W., Feng, X., Zhao, L., Li, Q., Huang, L., Gu, Y., Ma, W., Xu, Y., and Qin, B.
\newblock Investigating and mitigating the multimodal hallucination snowballing in large vision-language models.
\newblock In \emph{Proceedings of the 62nd Annual Meeting of the Association for Computational Linguistics (Volume 1: Long Papers)}, pp.\  11991--12011, 2024.

\bibitem[Zhou et~al.(2024)Zhou, Cui, Yoon, Zhang, Deng, Finn, Bansal, and Yao]{lure}
Zhou, Y., Cui, C., Yoon, J., Zhang, L., Deng, Z., Finn, C., Bansal, M., and Yao, H.
\newblock Analyzing and mitigating object hallucination in large vision-language models.
\newblock In \emph{The Twelfth International Conference on Learning Representations}, 2024.

\bibitem[Zhu et~al.(2023)Zhu, Chen, Shen, Li, and Elhoseiny]{minigpt4}
Zhu, D., Chen, J., Shen, X., Li, X., and Elhoseiny, M.
\newblock Minigpt-4: Enhancing vision-language understanding with advanced large language models.
\newblock \emph{arXiv preprint arXiv:2304.10592}, 2023.

\end{thebibliography}
\bibliographystyle{icml2025}


\newpage
\appendix
\onecolumn

\section{Human Evaluation Dataset Construction}\label{app:human_eval_data}
The results in Table \ref{tab:human_eval} are obtained through the following process:
\begin{enumerate}[leftmargin=*, nolistsep]
    \item Captions are generated for 100 DOCCI test images using LLaVA-v1.5-7B and InstructBLIP.
    \item Human labelers evaluate the captions from LLaVA-v1.5-7B and InstructBLIP for each image in terms of factuality.
    \item Caption pairs with similar factuality quality are excluded.
    \item For the remaining pairs, the correlation between human decisions and those made by each automated metric is measured.
\end{enumerate}
\section{Undesirable Bias in FACTSCORE}\label{app:human_eval_critical}
Metrics that rely solely on unimodal information are inherently susceptible to undesirable biases. For instance, metrics like FACTSCORE, which depend exclusively on reference captions, introduce stylistic bias tied to the specific style, tone, or phrasing of the references, unfairly favoring or penalizing captions based on these factors. To demonstrate this, we compare FACTSCORE with our factuality metric using human-labeled captions that are hallucination-free but stylistically different from DOCCI captions (HUMAN) \citep{iiw}. Table \ref{tab:app_human_eval} shows that, due to its stylistic bias, FACTSCORE assigns lower scores to these human-labeled captions, even though they are clearly superior to LLaVA-v1.5-7B and InstructBLIP captions in terms of factuality. In contrast, our factuality metric remains robust against such bias.\begin{table}[h]
\centering
\caption{Comparison of correlations between human preferences and automated metrics in terms of factuality.}
{
\begin{tabular*}{0.65\columnwidth}{@{\extracolsep{\fill}}ccc@{}}
\toprule
\multirow{2}{*}{Task} & \multicolumn{2}{c}{Spearman's $\rho$}\\
\cmidrule{2-3}
& FACTSCORE & Ours\\
\midrule[.05em]
LLaVA-v1.5-7B vs. InstructBLIP & 67.9 & 70.2\\
HUMAN vs. LLaVA-v1.5-7B vs. InstructBLIP & 18.3 & 61.4\\
\midrule[.1em]

\end{tabular*}
}
\label{tab:app_human_eval}
\end{table}
\section{Ablation Study}\label{app:ablation_study}
\begin{table}[h]
\centering
    \caption{Effectiveness of our proposed method across various captioning models as a function of $\pi$. In the CapMAS column, the LLM represents the decomposer and corrector, while the MLLM represents the fact-checker.}
{
\footnotesize
\begin{tabular*}{1.0\columnwidth}{@{\extracolsep{\fill}}ccccccc@{}}
\toprule
\multirow{2}{*}{Captioner} & \multicolumn{3}{c}{CapMAS} & \multicolumn{3}{c}{Metric}\\
\cmidrule(lr){2-4}\cmidrule(lr){5-7}
& LLM & MLLM & $\pi$ & CLAIR & Factuality & Coverage\\\midrule[.05em]
\multirow{4}{*}{LLaVA-NeXT-7B} & - & - & - & 68.8 &	59.9 &	47.9\\
& LLaMA-3-8B & LLaVA-NeXT-7B & 1.0 & 74.1 & 72.2 & 46.9\\
& LLaMA-3-8B & LLaVA-NeXT-7B & 0.5 & 73.6 & 76.9 & 43.7\\
& LLaMA-3-8B & LLaVA-NeXT-7B & 0.3 & 72.2 & 76.8 & 40.0\\
\midrule[.05em]
\multirow{4}{*}{LLaVA-NeXT-13B} & - & - & - & 70.2 & 62.1 & 48.5\\
& LLaMA-3-8B & LLaVA-NeXT-13B & 1.0 & 75.5 & 77.9 & 45.8\\
& LLaMA-3-8B & LLaVA-NeXT-13B & 0.5 & 74.8 & 79.9 & 42.1\\
& LLaMA-3-8B & LLaVA-NeXT-13B & 0.3 & 72.6 & 80.5 & 39.6\\
\midrule[.05em]
\multirow{4}{*}{InternVL-Chat-V1.5} & - & - & - & 74.9 & 65.5 & 48.2\\
& LLaMA-3-8B & InternVL-Chat-V1.5 & 1.0 & 78.2 & 75.9 & 47.3\\
& LLaMA-3-8B & InternVL-Chat-V1.5 & 0.5 & 79.0 & 78.8 & 46.0\\
& LLaMA-3-8B & InternVL-Chat-V1.5 & 0.3 & 77.7 & 81.7 & 42.5\\
\bottomrule

\end{tabular*}
}
\label{tab:appendix_pi}
\end{table}
Our proposed method features a single hyperparameter, $\pi$, which serves as the threshold for classifying atomic propositions as hallucinations or non-hallucinations. Table \ref{tab:appendix_pi} presents the effects of CapMAS across various models as a function of $\pi$. The results reveal a loose trade-off between factuality and coverage depending on $\pi$. Specifically, in all tested settings, as $\pi$ increases, factuality tends to decrease while coverage increases.
\newpage
\section{The complete version of Table \ref{tab:metaeval}}\label{app:metaeval_full}
\begin{table}[h]
\centering
\caption{Meta-evaluation results across various caption evaluation methods. DOCCI and its synthetic hallucinatory captions are used for the meta-evaluation. The highest-rated caption for each method is highlighted in \textbf{bold}.}
{
\footnotesize
\addtolength{\tabcolsep}{-2pt}
\begin{tabular*}{1.0\columnwidth}{@{\extracolsep{\fill}}cccccccccc@{}}
\toprule
\multirow{2}{*}{Caption} & \multicolumn{9}{c}{Evaluation Metric}\\
& BLEU & ROUGE & METEOR & CIDEr & CLIP-S & RefCLIP-S& CLAIR & ALOHa & Ours\\
\midrule[.05em]
Clean & 4.2 & 22.0 & 13.7 & 6.4 & 81.3 & 75.5 & \textbf{86.9} & 36.2& \textbf{62.8}\\
Object & \textbf{4.9} & \textbf{22.3} & \textbf{14.5} & 4.8 & 81.0 & 75.3 & 85.2 & 31.5& 52.3\\
Attribution & 4.1 & 21.8 & 13.6 & 6.2 & 80.9 & 75.2 & 80.0 & 34.3& 60.9\\
Relation & 4.1 & 21.8 & 13.7 & \textbf{6.7} & \textbf{81.4} & \textbf{75.6} & 83.5 & \textbf{36.9}& 51.9\\
\midrule[.1em]

\end{tabular*}
}
\label{tab:appen_metaeval}
\end{table}
\section{Case}
\begin{figure*}[h]
{
\begin{center}
\centerline{\includegraphics[width=1.0\textwidth]{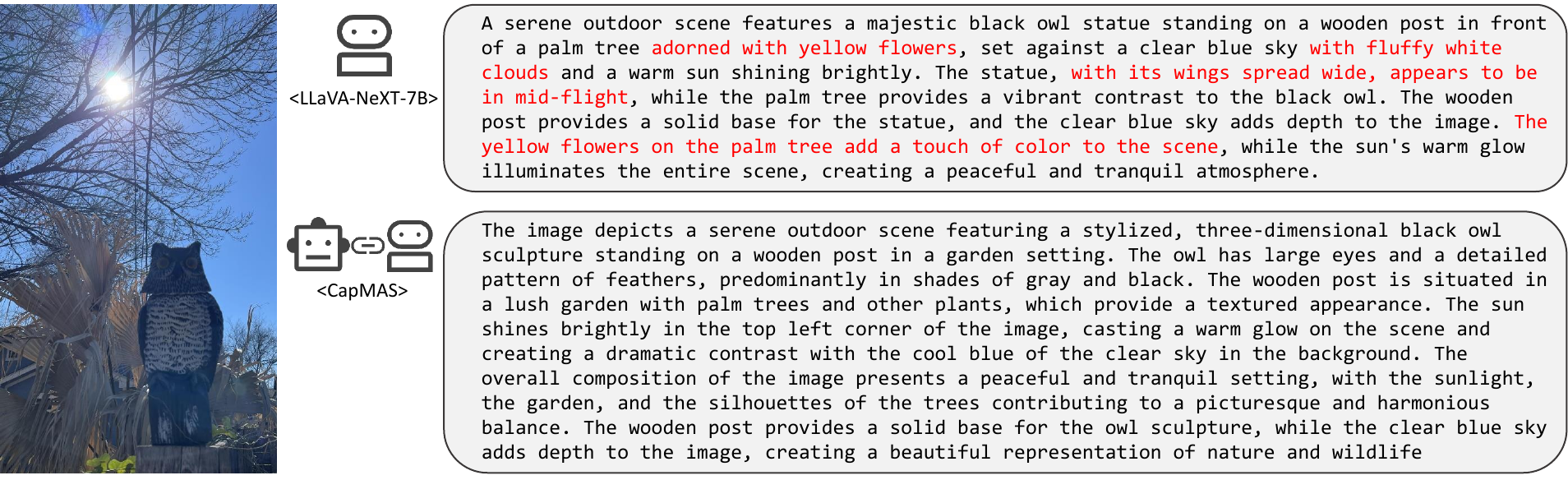}}%
\caption{An example of a caption generated by CapMAS, with LLaVA-NeXT-7B as both the captioning and fact-checking model and LLaMA-3-8B as both the decomposer and corrector LLM.}
\label{fig:case}
\end{center}
}
\end{figure*}
\section{Additional Experiments}\label{app:addexp}
\subsection{Fluency Analysis}
We investigate the effect of our framework on the fluency of the generated captions.
We employ an LLM-based evaluation to measure the basic quality of language in the captions generated by CapMAS. Specifically, we utilize GPT-4o with the following prompt:
\begin{quote}
You are a language expert evaluating the fluency of image captions.

Fluency refers to how grammatically correct, natural, and well-formed the text sounds to a native English speaker. A fluent caption should be grammatically correct, free of awkward phrasing, and read smoothly.

Evaluate the fluency of the following caption and return your output **strictly in JSON format** with:\newline
- ``reason'': a key reason for your scoring\newline
- ``score'': a number between 0 (completely disfluent) and 100 (perfect fluency)

Caption: ``\{caption\}''
\end{quote}

In addition to evaluating the captions generated by CapMAS, we also assess human-written, detailed captions for the same set of images.
\begin{table}[]
\centering
\caption{Fluency evaluation results. The results of CapMAS (captioner) demonstrate that the proposed framework can generate captions with higher fluency than the human-written reference captions (Human).}\label{tab:app:fluency}
{
\begin{tabular}{c @{\extracolsep{\fill}} c}
Captions generated by & Fluency $\uparrow$\\\midrule[.05em]
Human & 89.0\\
CapMAS (LLaVA-v1.5-7B)&93.4\\
CapMAS (LLaVA-NeXT-7B)&93.4\\
CapMAS (LLaVA-NeXT-13B)&93.6\\
CapMAS (InternVL-Chat-V1.5)&94.1\\
\midrule[.1em]
\end{tabular}
}
\end{table}

The results in Table \ref{tab:app:fluency} demonstrate that the captions generated by CapMAS achieve even higher fluency scores than the human-generated captions. This can be attributed to the final stage of CapMAS, in which the corrector LLM helps preserve or even improve the fluency of the captions.

\subsection{Effect of Self-Refine on Image Captioning}
CapMAS can be interpreted as an inference-time scaling strategy. However, allocating more compute at inference time does not necessarily lead to better results. To demonstrate this, we test the effectiveness of inference-time scaling through Self-Refine (SR) \cite{self_refine}. Since validating and revising one's output requires advanced reasoning capabilities, we conducted the experiments using GPT-4V. For SR, we used the following prompt:
\begin{quote}
You are given an image and its corresponding caption.
Your task is to:

1. Analyze the image and compare it with the caption.\newline
2. Identify and correct any factual or descriptive errors in the caption based on the image.\newline
3. Refine the caption for clarity, correctness, and completeness — even if the original caption is mostly accurate.

Show your reasoning and then provide a final refined caption.
\end{quote}

\begin{table}[]
\centering
\caption{Effect of Self-Refine (SR) on image captioning. The results show that the quality of captions degrades as SR iterations increase.}\label{tab:app:scaling}
{
\begin{tabular}{c @{\extracolsep{\fill}} ccc}
Method&CLAIR&Factuality&Coverage\\\midrule[.05em]
Base&82.4&77.1&53.5\\
 +SR (x1)&79.9&72.3&50.4\\
 +SR (x2)&79.1&70.6&50.1\\
 +SR (x3)&78.5&69.8&49.7\\\midrule[.1em]
\end{tabular}
}
\end{table}

Table \ref{tab:app:scaling} shows that having the model revise its own captions iteratively does not lead to better results.

\subsection{Efficiency of CapMAS}
CapMAS is an approach that achieves better results by incurring additional cost at inference time. Although it involves a multi-model pipeline, CapMAS can offer a better cost-performance trade-off than Self-Refine (SR) methods for the following reasons:
\begin{itemize}
    \item Most of CapMAS’s cost lies in the final step, where the corrector LLM generates a refined caption based on the initial caption and the True/False classification results of the propositions. The decomposition step involves shorter sequences, and the MLLM used for proposition classification can process them in parallel, generating only a single token (True or False) per proposition. SR processes long sequences that include the original caption, detailed feedback, and the refined caption. Considering the length and complexity of the feedback \cite{self_refine}, CapMAS and SR can be seen as comparable in cost.
    \item CapMAS’s performance does not heavily depend on the capability of the LLM used. This suggests that the LLM-related cost in the pipeline could potentially be further reduced.
\end{itemize}
\section{Prompt Templates}\label{app:prompts}
\begin{figure}[h]
{
\begin{center}
\centerline{\includegraphics[width=1.0\columnwidth]{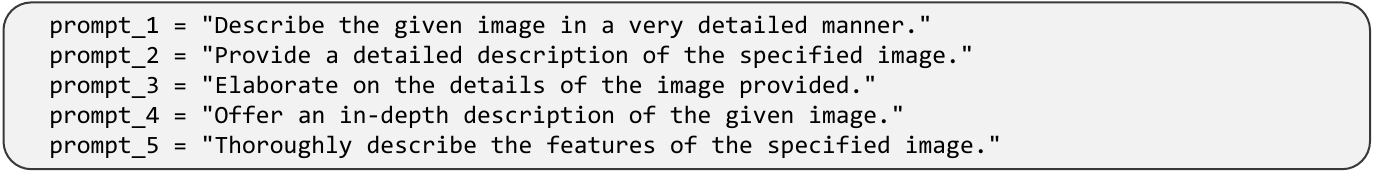}}%
\caption{The five prompt inputs used to generate captions in our experiments.}
\label{fig:appendix_caption}
\end{center}
}
\end{figure}
\begin{figure}[h]
{
\begin{center}
\centerline{\includegraphics[width=1.0\columnwidth]{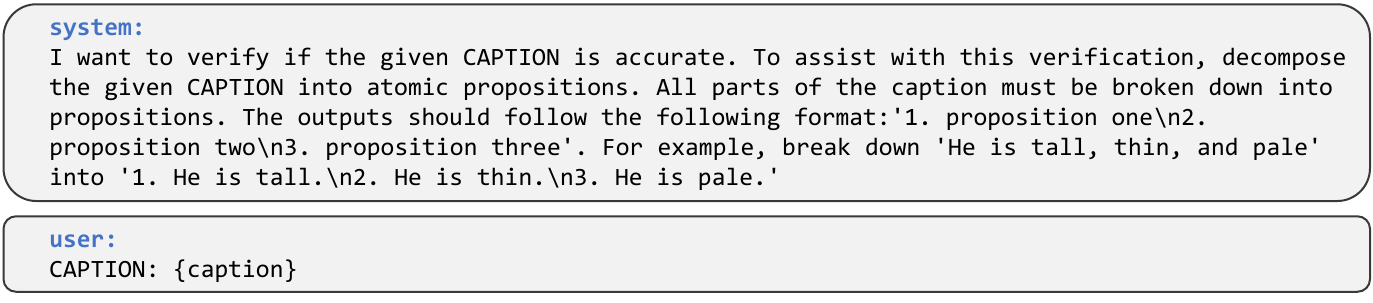}}%
\caption{The prompt input for LLaMA-3-8B serving as the decomposer.}
\label{fig:appendix_decompose}
\end{center}
}
\end{figure}
\begin{figure}[h]
{
\begin{center}
\centerline{\includegraphics[width=1.0\columnwidth]{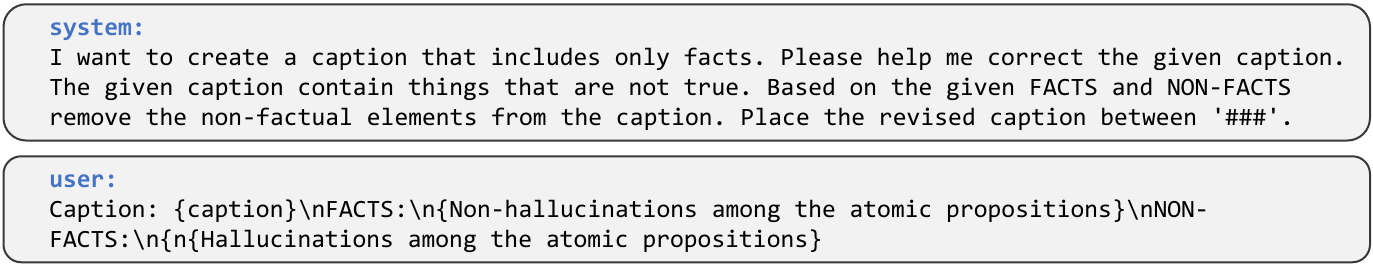}}%
\caption{The prompt input for LLaMA-3-8B serving as the corrector.}
\label{fig:appendix_correct}
\end{center}
}
\end{figure}
\begin{figure}[h]
{
\begin{center}
\centerline{\includegraphics[width=1.0\columnwidth]{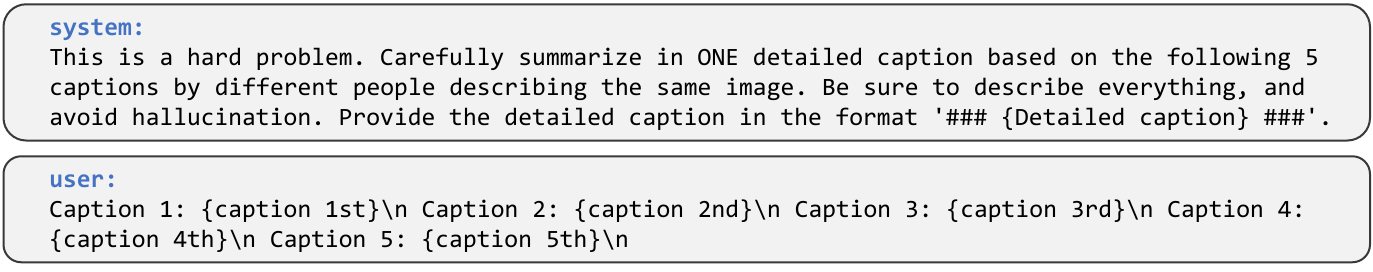}}%
\caption{The prompt input for LLaMA-3-8B serving as the summerizer. We use the prompt employed in the work of \cite{vfc}.}
\label{fig:appendix_decomposesummarize}
\end{center}
}
\end{figure}
\begin{figure}[h]
{
\begin{center}
\centerline{\includegraphics[width=1.0\columnwidth]{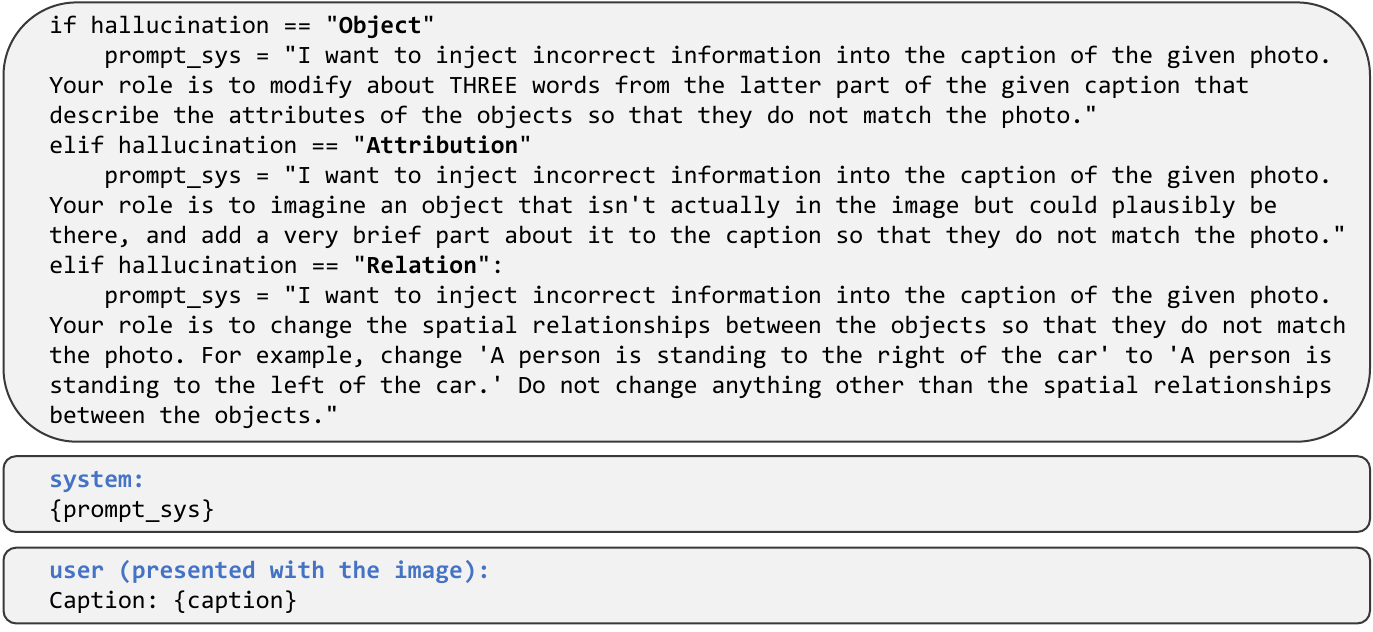}}%
\caption{The prompt input for GPT-4o used to create the meta-evaluation dataset of Table \ref{tab:metaeval}.}
\label{fig:appendix_metaeval}
\end{center}
}
\end{figure}
\begin{figure}[h]
{
\begin{center}
\centerline{\includegraphics[width=1.0\columnwidth]{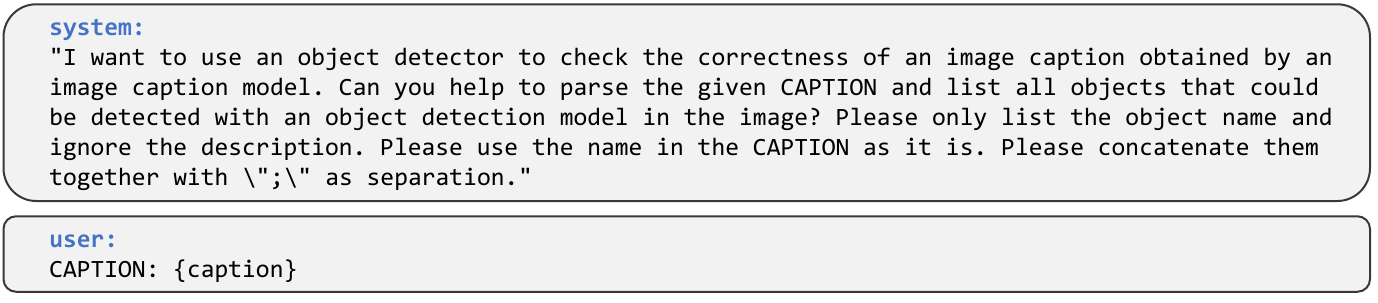}}%
\caption{The prompt input for GPT-4 used to create the dataset of Figure \ref{fig:motivation_dataset}. We use the prompt employed in the work of \cite{vfc}.}
\end{center}
}
\end{figure}

\end{document}